\documentclass[runningheads]{llncs}
\usepackage{graphicx}

\usepackage{tikz}
\usepackage{comment}
\usepackage{amsmath,amssymb} 
\usepackage{color}

\usepackage[accsupp]{axessibility}  

\usepackage{bm}  
\usepackage{xcolor} 
\usepackage{array}
\usepackage{xspace}

\usepackage{booktabs} 
\usepackage{multirow}
\usepackage{pdfpages}

\definecolor{myblue}{rgb}{0.4235294117647059,0.5568627450980392, 0.7490196078431373}
\definecolor{myred}{rgb}{0.7215686274509804,0.32941176470588235, 0.3137254901960784}

\usepackage[capitalize]{cleveref}
\crefname{section}{Sec.}{Secs.}
\Crefname{section}{Section}{Sections}
\Crefname{table}{Table}{Tables}
\crefname{table}{Tab.}{Tabs.}

\def\tabspacetop{{\vspace{-3mm}}}
\def\tabspacebot{{\vspace{-1mm}}}

\newcommand{\ch}{\small{ \mbox{\fontfamily{qcs}\selectfont{ch}}}}
\newcommand{\BS}{\small{ \mbox{\fontfamily{qcs}\selectfont{BS}}}}
\newcommand{\lr}{\small{ \mbox{\fontfamily{qcs}\selectfont{lr}}}}

\newif\ifrevfinal
\revfinalfalse
\def\rev[#1][#2]{\ifrevfinal #2 \else {\color{blue} \sout{#1}} {\bf \color{red} #2} \fi}
\usepackage{color}
\usepackage[normalem]{ulem}
\usepackage{amsmath,amssymb,amsbsy,xspace}

\def\R{{\rm I\!R}}                            	

\usepackage{amsmath}

\def\<{\langle}
\def\>{\rangle}

\def\Eq#1{Eq.~(\ref{eq:#1})}

\def\be{\begin{equation}}
\def\ee{\end{equation}}
\def\bea{\begin{eqnarray}}
\def\eea{\end{eqnarray}}

\def\tab#1{Table~\ref{tab:#1}}

\makeatletter
\gdef\SetFigFont#1#2#3{%
  \reset@font\fontsize{10}{12pt}%
  \selectfont%
}
\let\eqnarray@=\eqnarray \let\endeqnarray@=\endeqnarray
\def\eqnarray{\bgroup\arraycolsep=2pt\eqnarray@}
\def\endeqnarray{\endeqnarray@\egroup}
\makeatother
\def\arraystretch{1.2}
\def\em{\slshape}

\makeatletter
\DeclareRobustCommand\onedot{\futurelet\@let@token\@onedot}
\def\@onedot{\ifx\@let@token.\else.\null\fi\xspace}

\def\eg{\emph{e.g}\onedot} 
\def\ie{\emph{i.e}\onedot}

\def\wrt{w.r.t\onedot} 
\def\etal{\emph{et al}\onedot}

\newcounter{iictr}
\def\@iia[#1]{\setcounter{iictr}{#1}\@iib}
\def\@iib{\iii[\roman{iictr}]\xspace}
\def\ii{\@ifnextchar[{\@iia}{\stepcounter{iictr}\@iib}}
\def\iii[#1]{({\em#1\/})}
\makeatother

\def\mypar#1{\vspace{0mm}{\noindent\bf #1}\hspace{1mm}}

\def\wrt{w.r.t\onedot} 
\def\eg{\emph{e.g}\onedot} 

\begin{document}
\pagestyle{headings}
\mainmatter
\def\ECCVSubNumber{10}  

\title{Unifying conditional and unconditional\\ semantic image synthesis with \ours}  

\def\ours {{OCO-GAN}\xspace}

\titlerunning{Unifying conditional and unconditional  semantic image synthesis}
%
\author{Marl\`ene Careil\inst{1,2}\orcidID{0000-0003-4542-5371} 
\and
St\'ephane Lathuili\`ere\inst{2}\orcidID{0000-0001-6927-8930} 
\and
Camille Couprie\inst{1}
\and
Jakob Verbeek\inst{1}\orcidID{0000-0003-1419-1816}
}
\authorrunning{M. Careil et al.}
%
\institute{Meta AI \and
LTCI, T\'el\'ecom Paris, Institut Polytechnique de Paris}

\maketitle

\begin{abstract}

Generative image models have been extensively studied in recent years. In the unconditional setting, they  model the marginal distribution from unlabelled images. To allow for more control, image synthesis can be conditioned on semantic segmentation maps that instruct the generator the position of objects in the image. While these two tasks are intimately related, they are generally studied in isolation. 
We propose \ours, for Optionally COnditioned GAN, which addresses both tasks in a unified manner, with a shared image synthesis network that can be conditioned either on semantic maps or directly on latents. 
Trained adversarially in an end-to-end approach with a shared discriminator, we are able to leverage the synergy between both tasks.
We experiment with Cityscapes, COCO-Stuff, ADE20K datasets in a limited data, semi-supervised and full data regime and obtain excellent performance, improving over existing hybrid models that can generate both with and without conditioning in all settings. Moreover, our results are competitive  or better than  state-of-the art specialised unconditional and conditional models. 

\end{abstract}

\section{Introduction}

Remarkable progress has been made in modeling complex data distributions with deep  generative models, allowing considerable improvement in generative models for images~\cite{brock19iclr,ho20neurips,Karras2019stylegan2,ramesh21arxiv,vahdat20arxiv}, videos~\cite{walker21arxiv} and other contents~\cite{nash20icml,DBLP:journals/corr/YangXCLZHL17}. 
In particular, GANs~\cite{goodfellow14nips,karras20nips} stand out for their compelling sample quality, and are able to produce near-photo realistic images in restricted domains such as human faces~\cite{Karras2019stylegan2} or object centric images~\cite{brock19iclr,sauer22arxiv}.

Despite these advances, modelling  more complex indoor or outdoor scenes with many objects remains challenging~\cite{casanova21nips}.
To move beyond this limitation, and  allow for more control on the generation process,
conditional image generation based on semantic  segmentation maps or labeled bounding boxes has been investigated, see \eg ~\cite{park19cvpr1,sun2020learning,sushko21iclr,sylvain2020object}. 
Moreover, semantically conditioned image generation, or ``semantic image synthesis'' for short, 
enables users  with precise control over what content should be generated where in automated content generation. Yet, training semantic image generation models requires images annotated with  segmentation maps, which are expensive to acquire due to their detailed nature.

Unconditional and conditional image generation are closely related, and  indeed to some extent semantics can be recovered from unconditionally trained GANs, see \eg ~\cite{bau19iclr,voynov20icml}.
Despite their relatedness, however, both tasks are generally studied in isolation, even though they could potentially mutually benefit each other. 
Unlabeled images do not require expensive annotation, and jointly training an unconditional generator may help conditional generation since parameter-sharing acts as a form of regularization. 
In particular in semi-supervised settings where only few images are annotated, jointly training a single hybrid model can help conditional generation.
Conversely, annotated images can provide per-pixel loss terms to train the network, and help the unconditional network to output images with better structure and higher levels of detail. 
This supervision, can be especially useful in settings with small training datasets where unconditional generation is challenging.

\begin{figure}[t]
\begin{minipage}{0.59\linewidth}
    \includegraphics[width=\linewidth]{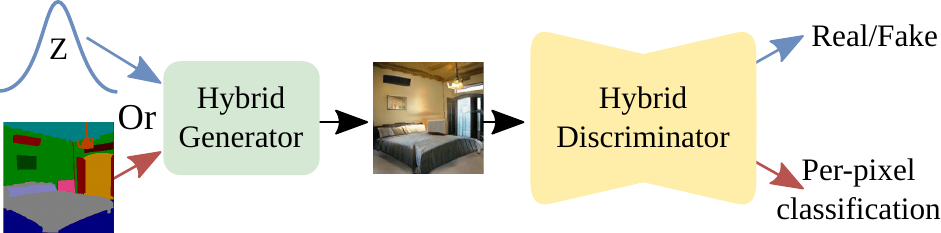}
\end{minipage}
\hfill
\begin{minipage}{0.4\linewidth}
\centering
\scriptsize
\def\myim#1{ \includegraphics[width=10mm]{figures/#1}}
    { \textcolor{myblue}{Unconditional generation}}\\
    \myim{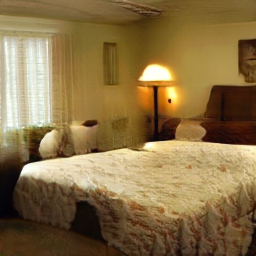}
    \myim{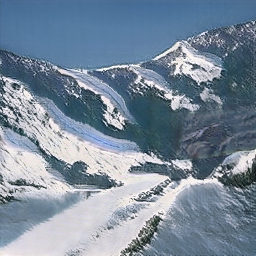}
    \myim{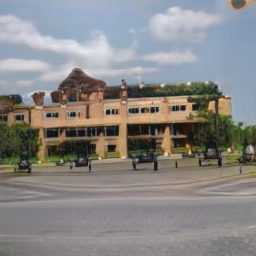}
    \myim{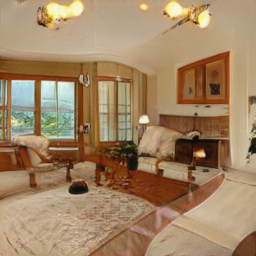}\\
   {\textcolor{myred}{Conditional generation}}\\
\myim{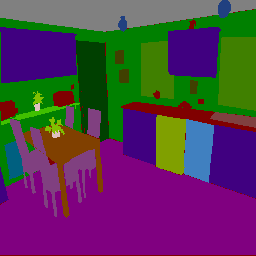}
\myim{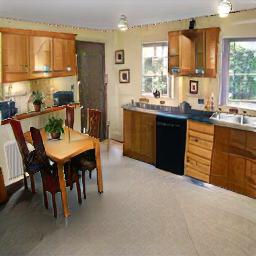}
\myim{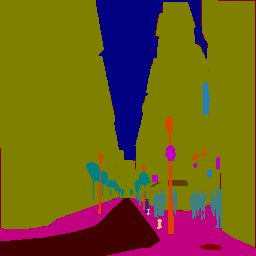}
\myim{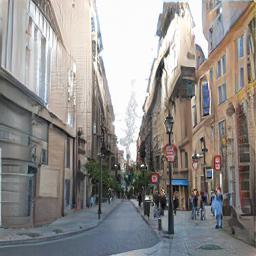}
\end{minipage}
    \caption{ Our \ours uses
    a shared synthesis network that is either conditioned on a semantic segmentation map, 
    or on latent variables only for unconditional generation.  
    The shared discriminator provides a training signal by classifying entire images as real or fake, and uses an additional per-pixel classification loss for conditionally generated images. 
    We show 
    image generation examples with \ours trained on ADE20K.
}
    \label{fig:teaser}
\end{figure}

The few works that address both problems together, take an existing conditional or unconditional model, and add a module to address the other task.
For example, by taking a pre-trained unconditional GAN, and learning an inference network that takes a semantic map as input and which produces the GAN latents that reconstruct the corresponding training image~\cite{richardson21cvpr}.
Another approach is to learn a semantic map generating network, which can be combined with  a conditional semantic synthesis network to perform unconditional generation~\cite{azadi2019semantic}.
Such stage-wise approaches, however, do not fully leverage the potential of learning both tasks simultaneously. In the former approach the unconditional GAN is pre-trained without taking advantage of the semantic segmentation data, while the latter does not use unlabeled images to train the image synthesis network.

To address limited data and semi-supervised  settings,  we introduce \ours, for Optionally COnditioned GAN, which enables both unconditional generation as well as conditional semantic image synthesis in a single framework, where most of the parameters between the two tasks are shared. 
We also propose a single end-to-end training process that does not require stage-wise training like previous hybrid models resulting in performance gain for unconditional and semantic image synthesis.
See Figure \ref{fig:teaser} for an  illustration. 
Building off StyleMapGAN~\cite{kim21cvpr},  our model uses a style-based synthesis network with locally defined styles.
The  synthesis network  is shared across both tasks, and is either conditioned on a semantic map or on latent variables. 
To train the model, we use a discriminator that is also shared between both tasks. 
A real/fake classification head for the entire image is used to train for unconditional generation, and  a per-pixel classification loss~\cite{sushko21iclr} is used for  conditional  generation.

We perform experiments on the Cityscapes, COCO-Stuff and ADE20K datasets for the two addressed settings: for the limited data regime  we train the models on small subsets of the original datasets,  for the semi-supervised setting  we train from the full datasets but use labels only for a subset of the images.
To complete our evaluation, we also report results of experiments using the full datasets.

In summary, our contributions are the following: 
\begin{itemize}
\setlength\itemsep{0em}
\item 
We propose \ours, a unified style-based model capable of both unconditional image generation as well as conditional semantic image generation. 
\item 
We introduce an end-to-end training procedure that combines conditional and unconditional losses using a shared discriminator network.
\item Our experiments show that OCO-GAN improves over existing state-of-the-art hybrid approaches, while also yielding results that are competitive or better than those of state-of-the-art specialized methods for either conditional or unconditional generation in the three addressed settings.
\end{itemize}

\section{Related Work}
\label{sec:related}

\mypar{GANs for image generation without spatial guidance.}
GANs have been widely used in computer vision applications, due to their excellent performance at synthesizing (near-)photo-realistic images. State-of-the-art unconditional generative models include StyleGAN~\cite{Karras2019stylegan2,karras21nips,karras19cvpr} and BigGAN~\cite{brock19iclr}.  
StyleGAN2~\cite{Karras2019stylegan2} is  composed of two main components: a mapping network that produces the style of an image, and a synthesis network that incorporates the style at different feature levels through convolutional layers to generate the image. 
BigGAN uses a convolutional upsampling architecture, and conditions on class labels by using class-specific gains and biases in the BatchNorm layers.
Unconditional GANs, such as StyleGAN2, obtain very good results when trained on datasets with limited diversity,  such as ones containing only human faces, or a single class of objects such as cars.
For more diverse datasets, \eg  ImageNet,  class-conditional GANs such as BigGAN or StyleGAN-XL~\cite{sauer22arxiv} allow for better generation quality by relying on manually defined labels for training and generation.
StyleMapGAN~\cite{kim21cvpr} adds locality to the styles that are used in StyleGAN for affine  modulation of  the features in the layers of the image synthesis network.
In our work we build upon the StyleMapGAN architecture, as the spatiality of the styles  fits well with the semantic image generation task.

\mypar{GANs for semantic image synthesis.}
Conditional GANs for semantic image synthesis, a task consisting of generating photo-realistic images from semantic segmentation masks, have considerably developed over the last few years and offer more control over the generation process. 
Various models with different architectures and training techniques have been proposed. Pix2pix \cite{isola2017pix2pix} is one of the first works, and proposes to use a U-Net\cite{ronneberger15miccai} generator with a patch-based discriminator. SPADE~\cite{park19cvpr} introduces spatially adaptive normalization layers that modulate the intermediate features with the input labels, which efficiently propagate semantic condition through the generator, and uses a multi-scale patch-based  discriminator that evaluates the image/label pairs at different scales. CC-FPSE~\cite{liu19nips} also modifies the generator to produce conditional spatially-varying convolutional kernels and adopts a feature-pyramid discriminator. SC-GAN~\cite{wang2021image} divides the generator into two tasks: semantic encoding and style rendering. These semantic vector generator is trained with a regression loss, while  adversarial and perceptual losses are used to train the style rendering generator. 
CollageGAN~\cite{Li_2021_ICCV} aims at high resolution semantic image synthesis by developing a conditional version of StyleGAN2 and using class-specific generators, trained on additional datasets, which enable it to have better image quality for small objects. 
OASIS~\cite{sushko21iclr} adopts a new type of discriminator: a U-Net which is trained to predict the segmentation labels of pixels in real images and an additional ``fake'' label for pixels in generated images. 
In contrast to previous methods that are trained with  perceptual  and $\ell_2$ reconstruction losses, OASIS demonstrates good performance when using only the adversarial loss, showing better  diversity in the generated images without sacrificing image quality.
In our work we build upon the state-of-the-art OASIS architecture, extending it to enable both conditional and unconditional generation.

\mypar{Hybrid conditional-unconditional GANs.}
Almost all GANs in the literature are  trained for either
conditional or unconditional image synthesis. 
Existing hybrid approaches combine conditional and unconditional blocks in stage-wise training procedures.
For example they  train an encoder on top of a pre-trained unconditional GAN to perform conditional generation or image editing~\cite{nguyen2017plu,richardson21cvpr,zhou17cvpr,Zhu2020,zhu2016eccv}. 
Among them, pixel2style2pixel (PSP)~\cite{richardson21cvpr} is a network built for various tasks of image-to-image translation, that produces style vectors in the $W+$ latent space of  a pre-trained StyleGAN2 generator.  
Semantic Bottleneck GAN (SB-GAN)~\cite{azadi2019semantic} treats the semantic map as a latent variable to enable unconditional image synthesis. 
They separately train SPADE for conditional image generation, and a second unconditional  model that can generate  semantic segmentation maps. 
They then  fine-tuning these two networks in an end-to-end approach, using Gumbel-Softmax approximation to differentiate through the discrete segmentation maps.
They obtain better results than ProGAN \cite{karras18iclr} and BigGAN~\cite{brock19iclr} for unconditional generation, and SPADE for conditional synthesis on the Cityscapes and ADE-indoor datasets. In contrast to these approaches, our \ours is based on a single end-to-end architecture, which takes semantic maps or latents as conditioning input to the shared generator, and uses a shared discriminator with a per-pixel loss and entire image classification loss to obtain the training signal for conditional and unconditional generation respectively. 
Training our hybrid architecture in a single training process with mixed batches of conditional and unconditional samples allows us to better leverage the synergy between both tasks, as most of the network parameters are in the shared parts of the architecture, and are consistently trained by both losses.
Recently, \cite{huang2021poegan} introduced Product-Of-Experts GANs, able to synthesize images from multiple modalities, including text, segmentation maps and sketches,  by using a product of experts to model a hierarchical conditional latent space and training with a multimodal discriminator.
\mypar{Learning GANs from limited data.}
Improving GANs by discriminator augmentation has recently been explored in several works to regularize the discriminator in low-data regimes~\cite{karras20nips,tran20arxiv,zhao20nips,zhao20arxiv}.
For example,  \cite{karras20nips} uses an adaptive  augmentation method (ADA), applying augmentations on real and synthetic images with a certain probability before they enter the discriminator.
Specifically, the probability of augmentation is  set proportionally to the fraction of real images receiving positive discriminator output. 
With this approach,  the results of StyleGAN2 can often be matched using an order of magnitude less data.
Another approach is explored in  \cite{lucic19icml}, which combines self-supervised and semi-supervised training. A feature extractor is learned with two heads for rotation and class label prediction respectively. It labels remaining data with a few-shot classifier, and trains a conditional GAN. 
These augmentation techniques are orthogonal to our hybrid \ours approach, and we show experimentally that ADA can be successfully applied to \ours.

\begin{figure}[t]
\centering
\tabspacetop
\includegraphics[width=\linewidth]{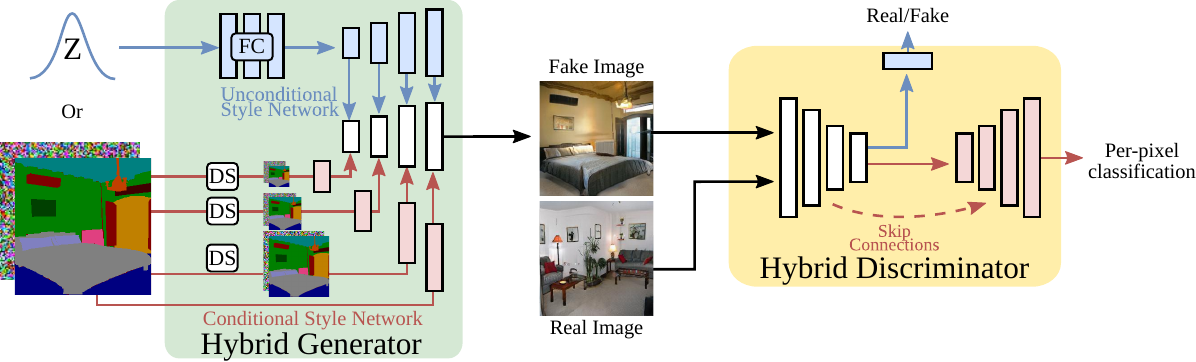}
\caption{Overview of \ours  which enables conditional (red path) and un-conditional (blue path) generation. 
Our hybrid generator takes alternatively a random vector, transformed into style maps in our unconditional style network in blue, or the concatenation of a segmentation map with a random map, going through down-sampling (DS) layers into our conditional style network in red.   
}
\tabspacebot
\label{overview}
\end{figure}

\section{Method}
\label{sec:method}

In this section we present \ours, which can generate images both in an unconditional manner, as well as conditioned on semantic segmentation maps.
For training  we leverage datasets composed of  RGB images ${\bf x}_n\in\R^{H\times W\times 3}, n\in\{1,\dots,N\}$. 
For part or all of the images, we have corresponding segmentation maps ${\bf s}_n\in\{0,1\}^{H\times W\times C}$, represented with one-hot encoding across $C$  classes. \ours  consists of a shared synthesis network 
which takes input from either a conditional or unconditional style network. 
To train our hybrid model, we take an adversarial approach and use a shared discriminator network with separate branches for conditional and unconditional generation. An overview of the full pipeline is displayed in Figure~\ref{overview}, and we provide more details in the following.

\subsection{Hybrid \ours architecture}
\label{sec:archi}

\mypar{Generator.} Inspired by recent success of style-based architectures~\cite{karras19cvpr,Karras2019stylegan2,sauer22arxiv}, we rely on  a style space that is shared between the conditional and unconditional generation tasks. 
The  generator is composed of two style networks corresponding to the conditional and unconditional cases. 
Their style representations are then processed by the shared image synthesis network. 
Sharing parameters between the conditional and unconditional tasks helps the unconditional branch to benefit from the available annotations in the limited data setting, while taking advantage of the large  available quantity of unlabeled images in the partially annotated setting. 
We adopt a style representation with spatial dimensions, which captures  more local variation  and is better suited for semantic image synthesis. 

The {\bf unconditional style network} $S_U$ is inspired from StyleMapGAN~\cite{kim21cvpr} which introduces spatial dimensions into style-based architectures. It consists of an MLP  mapping network that transforms the latent variable ${\bf z}\sim {\mathcal N}({\bf 0}, {\bf I})$  into a style map, which is transformed at different resolutions by a stack of convolutional and up-sampling layers, to match the size of feature maps in the synthesis network.

The {\bf conditional style network} $S_C$ takes a  segmentation map as input  and  produces spatial styles at different resolutions to feed into the synthesis network. 
To allow for diverse generations for a given semantic map, we follow OASIS~\cite{sushko21iclr} and concatenate a 3D noise tensor  ${\bf Z} \in  \mathbb{R}^{ H\times W\times 64}$  to the input semantic map along the channel dimension.
The noise tensor is obtained by spatially replicating a noise vector sampled from a unit Gaussian  of dimension 64.
We  resize the extended input semantic map and apply several convolutional layers, to produce the style maps at  matching resolutions for the layers of the synthesis network. 

The styles from the conditional or unconditional style network are then incorporated in the shared  {\bf synthesis network} $S_G$ through modulated convolutional layers  in every residual block. 
The activations in the synthesis network are first normalized  by subtracting and dividing by the mean and standard deviation computed  per channel and per image.  
The styles are then convolved to generate gains  and biases that are respectively multiplied with and added to the normalized feature maps~\cite{karras19cvpr}. 
We progressively upsample the features through transposed convolutions in residual blocks.

\mypar{Discriminator.}
For our discriminator D, we use a U-Net~\cite{ronneberger15miccai} encoder-decoder architecture with skip connections, based on the one from  OASIS~\cite{sushko21iclr}.
The decoder  outputs a per-pixel classification map across $C\!+\!1$ classes, representing the  $C$ labels in the dataset and an additional ``fake'' label.
The task of the discriminator consists in segmenting real images using supervision from the ground truth label maps and classifying pixels in generated images as fake. 
The generator, on the other hand, aims to make the discriminator recognize generated pixels as belonging to the corresponding class in the ground truth label map.

Our discriminator differs from the one in OASIS in several respects. 
First, we add a classification head to the output of the encoder that consists of convolutional and fully connected layers, and which is used to map the entire input image to a real/fake classification score.  The parameters in the discriminator's encoder are shared between both tasks to favor synergy during training. Parameter sharing is particularly useful in the limited data and partially annotated settings where each task can benefit from the other. Second, unlike OASIS, we don't use spectral normalization in the encoder part, but we find it beneficial to keep it in the decoder network. 
Third, in the decoder, before applying the residual upsampling blocks, we transform the output of the encoder by an {\it atrous} spatial pyramid pooling~\cite{DBLP:journals/corr/ChenPSA17} module which has multiple {\it atrous} convolutions in parallel at different rates. As it increases the spacing between the convolution kernel elements, {\it atrous} convolutions allow to increase receptive fields while maintaining spatial dimension, making them well suited for segmentation tasks~\cite{chen2018encoderdecoder}. 

Details on the network architectures are given in the supplementary material.

\subsection{Training}
\label{sec:training}

We train our models for both tasks jointly using the two discriminator branches, 
and use mixed batches of conditional and unconditional samples in training. 

The unconditional discriminator branch classifies entire images as real or fake, and is trained using the standard binary cross-entropy loss:
\begin{equation}
    \mathcal{L}_D^u = -\mathbb{E}_{{\bf x}}[\log D_{enc}({\bf  x})] - \mathbb{E}_{{\bf z}}[\log (1-D_{enc}(S_{G,U}({\bf z}))],
\end{equation}
where $S_{G,U}({\bf z})= S_G(S_U({\bf z}))$,
the expectation over ${\bf x}$ is with respect to the  empirical distribution of training images, and the expectation over ${\bf z}$  with respect to its unit Gaussian prior. 
We use the non-saturating GAN loss~\cite{goodfellow14nips} to train the unconditional branch of the generator network:
\begin{equation}
    \mathcal{L}_G^u = -\mathbb{E}_{z}[\log D_{enc}(S_{G,U}({\bf z}))].
\end{equation}
In  preliminary experiments, we observed best performance when this loss is applied only for  unconditional generation.

For the conditional discriminator branch, we rely on the multi-class cross-entropy loss of OASIS~\cite{sushko21iclr}, which aims to classify pixels from real images according to their ground-truth class, and pixels in generated images as fake:
\begin{eqnarray}
\nonumber
    \mathcal{L}_D^c &= &-\mathbb{E}_{({\bf x,s})}\left[\sum_{k=1}^C \alpha_k \sum_{h,w}^{H\times W}{\bf s}_{h,w,k} \log D_{dec}({\bf  x})_{h,w,k}\right] \\
    && -\mathbb{E}_{({\bf z,s})}\left[ \sum_{h,w}^{H\times W} \log D_{dec}(S_{G,C}({\bf z,s}))_{h,w,C+1}\right],
    \label{eq:CondDiscLoss}
\end{eqnarray}
where the $\alpha_k$ are class balancing terms, set to the inverse class frequency, 
expectation over ({\bf x,s}) is \wrt the empirical distribution, and over ({\bf z,s}) is \wrt the empirical distribution of segmentation maps $\bf s$ in the train set and the prior over $\bf z$.
As for the generator, we use the non-saturating version of \Eq{CondDiscLoss}:
\begin{equation}
\nonumber
  \mathcal{L}_G^c\!=\!-\!\mathbb{E}_{({\bf z,s})}\!\left[\sum_{k,h,w=1}^{C,H,W}\!\alpha_k 
  {\bf s}_{h,w,k} \log D_{dec}(S_{G,C}({\bf z,s}))_{h,w,C+1}\right]\!, 
    \label{eq:GenDiscLoss}  
\end{equation}
which aims to make the discriminator mistake generated pixels for each class with real pixels of that class.

To further improve the training of our models, we use additional regularization terms on both discriminator branches.
For the unconditional branch, we add the R1 regularization loss \cite{DBLP:journals/corr/abs-1801-04406} also used in  StyleGAN2, that penalizes gradients of the discriminator on real images with high Euclidean norm.  
Like in StyleGAN2, we perform this regularization every 16 minibatches.
For the conditional branch, we add the LabelMix loss of~\cite{sushko21iclr}, which mixes a real training image with a generated one using the  segmentation map of the real image. The loss aims to minimize differences in the class logits computed from the mixed image and the mixed logits computed from the real and generated images.

\section{Experiments}
\label{sec:expe}

We first describe our experimental setup, and then present quantitative and qualitative  results in  different settings, followed by  ablation studies. 

\subsection{Experimental setup}

\mypar{Datasets.} To validate our models we use the  Cityscapes~\cite{cordts16cvpr}, COCO-Stuff~\cite{caesar18cvpr}, ADE20K~\cite{zhou17cvpr} and CelebA-HQ~\cite{celeba} datasets, and use the standard validation and training sets used in the GAN literature ~\cite{park19cvpr,sushko21iclr,casanova21nips}.
We resize images and segmentation maps to $256\!\times\!256$, except for Cityscapes where it is  $256\!\times\!512$.
Besides experiments on the full datasets, we consider the following two settings.

In the \emph{limited data setting}, we use smaller subsets of the four datasets
to test models. 
This is interesting as obtaining semantic maps for training is time consuming.
As in~\cite{azadi2019semantic}, we use Cityscapes5K to refer to the subset of the 25K images in the full dataset for which fine semantic segmentation maps are available.  
For COCO-Stuff we randomly selected a subset of 12K images (10\% of all images), which we refer to as COCO-Stuff12K. ADE-Indoor is a subset of ADE20K  used in~\cite{azadi2019semantic} consisting of   4,377 training and 433 validation samples of indoor scenes. 
Moreover, we sampled a random subset of 4K images (20\% of all data), which we denote ADE4K. For CelebA-HQ dataset, we train models using 3k images.

In a \emph{partially annotated setting}, we test to what extent using a large number of unlabeled images is beneficial in cases where limited labeled training images are available.
Here we train with the full datasets, but use only a small portion of the semantic maps.
We used only 500 labeled images for Cityscapes and CelebA-HQ datasets, and  2K labels for ADE20K and COCO-Stuff datasets. Due space limitations, we report results on CelebA-HQ in the supplementary material. 

We will release our data spits  along with our code and trained  models upon publication  to facilitate reproduction of our results.

\begin{table}[t]
\caption{
Results in the {\bf limited data setting}. In italic: Results  taken from original papers.
In bold: Best result among  unconditional, conditional, and hybrid models, resp.
}
\centering
\tabspacetop
 \setlength{\tabcolsep}{1pt} 
\resizebox{\textwidth}{!}
{\scriptsize
\begin{tabular}{lrrr|rrr|rrr|rrr}
\toprule
  &  \multicolumn{3}{c|}{\textbf{Cityscapes5K}} &  \multicolumn{3}{c|}{\textbf{ADE-Indoor}} & \multicolumn{3}{c|}{\textbf{ADE4K}} & \multicolumn{3}{c}{\textbf{COCO-Stuff12K}}\\
 & 
 $\downarrow$\textbf{FID}  &   $\downarrow$\textbf{CFID} &    $\uparrow$\textbf{mIoU}   &
 $\downarrow$\textbf{FID}  &   $\downarrow$\textbf{CFID} &    $\uparrow$\textbf{mIoU}   &
 $\downarrow$\textbf{FID}  &   $\downarrow$\textbf{CFID} &    $\uparrow$\textbf{mIoU}   &
 $\downarrow$\textbf{FID}  &   $\downarrow$\textbf{CFID} &    $\uparrow$\textbf{mIoU}   \\  \midrule
StyleGAN2~\cite{Karras2019stylegan2} &  79.6 &   &  & 87.3 &   &  & 95.7  &  &  &  77.4 &  &   \\
StyleMapGAN~\cite{kim21cvpr} &  \bf 70.4  &   &  & \bf 74.8  & &  & \bf 87.8  &  &  &  \bf 65.2  &   &   \\
BigGAN~\cite{brock19iclr} &  80.5 &   &  & 109.6 & &  & 112.7 &  &  &  82.2  &   &   \\
\midrule

SPADE~\cite{park19cvpr1} &   & \it 71.8 & \it 62.3 &  & \bf 50.3  &  44.7 &  & {\bf 43.4}  & 34.3  &  & 28.3  & 33.8\\
OASIS~\cite{sushko21iclr} &   & \it \bf 47.7 & \it \bf 69.3 &  & 50.8  & \bf 57.5 &  & 49.9   & {\bf36.1} &  & {\bf 25.6} & \bf 37.6 \\
\midrule
PSP~\cite{richardson21cvpr} &  79.6  & 118.7   & 19.3 & 87.3  & 148.7 & 35.6 & 95.7  & 157.7  & 3.2 & 77.4  & 225.1 &    3.1 \\
SB-GAN~\cite{azadi2019semantic} & \it 65.5 &   \it 60.4  & n/a  & \it 85.3  & \it 48.2 &  n/a & 101.5  & 49.3  & 24.9 & 99.6 & 52.6 & 27.2 \\
 \ours (Ours) &  \bf 57.4 &  {\bf 41.6}& {\bf 69.4} & \bf 69.0 & \bf 47.9 &  \bf 58.6 & {\bf 80.6} & \bf  43.8   &  {\bf36.1} & {\bf 51.5}  & {\bf25.6}  & {\bf40.3}\\

\bottomrule
\end{tabular}
}
\tabspacebot
\label{tab:smalldatamain}
\end{table}

\mypar{Evaluation metrics.}
To assess image quality we report the standard {\bf FID} metric~\cite{heusel17nips} for both unconditional and conditional generation. 
We use {\bf CFID} to denote the FID metric computed for  {\bf C}onditionally generated images.
To assess the consistency of semantic image synthesis with the corresponding segmentation map, we report the {\bf mIoU} metric, calculated  using  the same segmentation networks as~\cite{sushko21iclr}.
We report  metrics averaged over five sets of samples, and provide standard   deviations and more evaluation details in the supplementary material.

\mypar{Baselines.}
We  compare to two hybrid models, capable of both conditional and unconditional image generation. 
To our knowledge, they are the only existing methods that handle both types of generation.
SB-GAN\cite{azadi2019semantic} 
was evaluated on Cityscapes5K, Cityscapes and ADE-Indoor, and  we train it on the other datasets using the settings provided by the authors for ADE-Indoor. 
PSP~\cite{richardson21cvpr} trains an encoder on top of a (fixed) pre-trained StyleGAN2 generator, and  was only evaluated on human face and animal face datasets.
As unconditional models we include StyleGAN2~\cite{Karras2019stylegan2}, StyleMAPGAN~\cite{kim21cvpr}, and BigGAN~\cite{brock19iclr}. 
For the class-conditional  BigGAN, we use a single trivial class label for all images to obtain an unconditional model, as in~\cite{azadi2019semantic,casanova21nips,lucic19icml,noroozi2020selflabeled}.
For semantic image synthesis we compare our method to the influential SPADE approach~\cite{park19cvpr1}, and the recent  OASIS~\cite{sushko21iclr} model which reports the best performance on the (full) datasets we consider.

Where possible we report results from the original papers for the baselines (in italic). In other cases  we train models using code released by the authors. In the partially annotated case, we train SB-GAN and PSP with the same  labelled and unlabelled images as  \ours. For PSP method, we use a pretrained StyleGAN2 on all the unconditional dataset and train the encoder using the limited amount of labelled data, while for SB-GAN, we train semantic synthesizer and SPADE generator with limited labeled data and use all the dataset for training the unconditional discriminator in the finetuning training phase.

\begin{figure}[t]
\caption{Unconditional (left, first three columns) and conditional (right, last three columns) generations from models in the {\bf limited data setting}, trained on ADE-Indoor (top three rows) and COCO-Stuff12K (bottom three rows).
SB-GAN unconditional images for ADE-Indoor taken from original paper.
}
\medskip
 \def\myim#1{ \includegraphics[width=16.5mm,height=16.5mm]{figures/#1}}
   \begin{minipage}[t]{0.37\textwidth}
   \setlength\tabcolsep{0.5pt}
   \renewcommand{\arraystretch}{0.2}
    \tiny
     \begin{tabular}{cccc}
         &StyleMapGAN & SB-GAN& \ours\\
\parbox[t]{2mm}{\multirow{3}{*}{\rotatebox[origin=c]{90}{ADE-Indoor}}} & \myim{/adeindoor_uncond/stylemapgan_bed.png} &
\myim{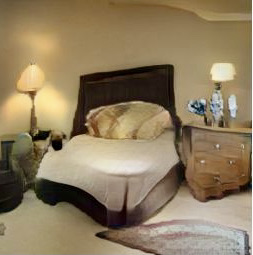} &
\myim{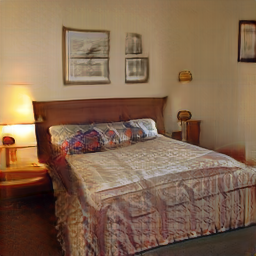}\\
&\myim{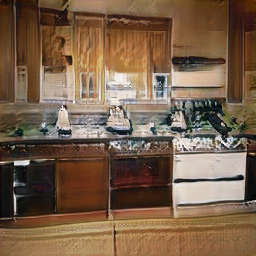} &
\myim{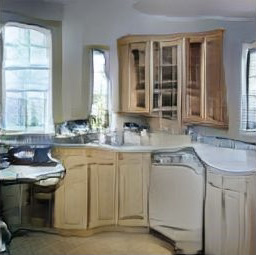} &
\myim{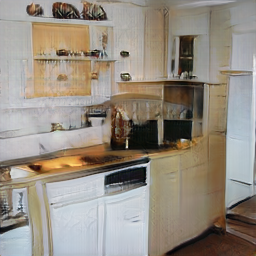} \\
&\myim{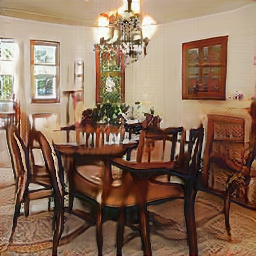} &
\myim{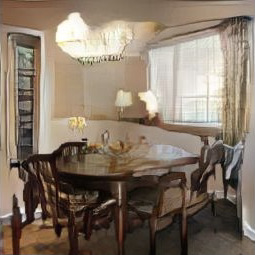} &
\myim{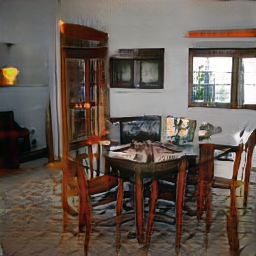} 
      \end{tabular}
   \end{minipage}
   \hspace{7mm}
   \begin{minipage}[t]{0.62\textwidth}
   \setlength\tabcolsep{0.5 pt}
   \renewcommand{\arraystretch}{0.2}
   \tiny
     \begin{tabular}{cccc}
    & OASIS& SB-GAN& \ours\\
\myim{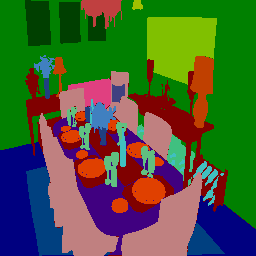} &
\myim{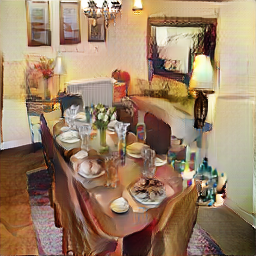} &
\myim{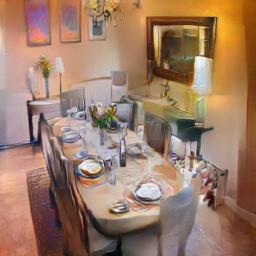}&
\myim{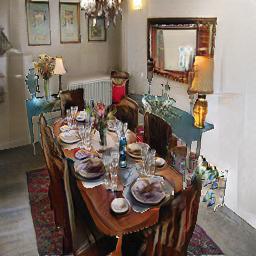}  \\
\myim{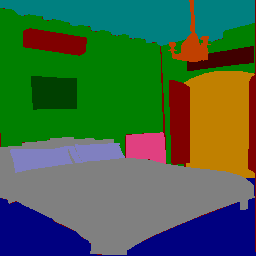} &
\myim{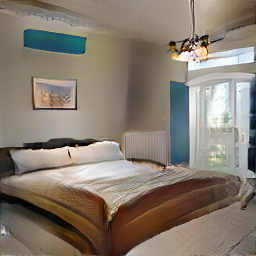} &
\myim{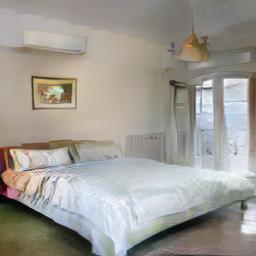}&
\myim{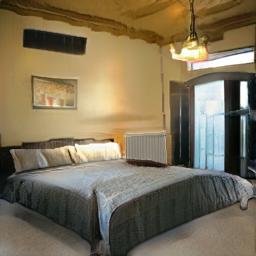}  \\
\myim{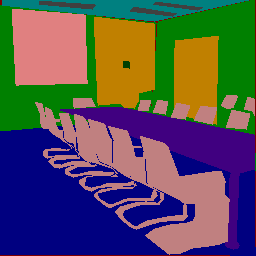} &
\myim{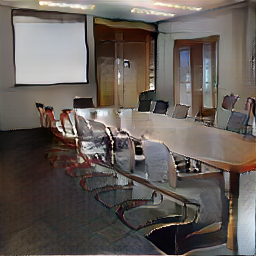} &
\myim{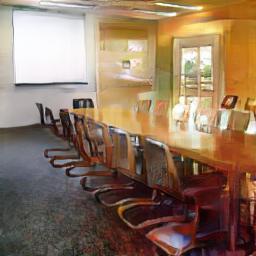}&
\myim{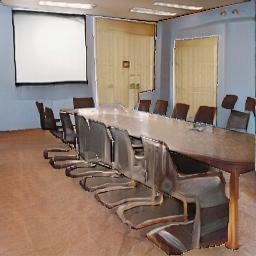}  \\
     \end{tabular}
\end{minipage} 

\vspace{1mm}
  \begin{minipage}[t]{0.37\textwidth}
  \setlength\tabcolsep{0.5 pt}
  \renewcommand{\arraystretch}{0.2}
    \tiny
     \begin{tabular}{cccc}
     &StyleMapGAN & SB-GAN& \ours\\
    \parbox[t]{2mm}{\multirow{3}{*}{\rotatebox[origin=c]{90}{COCO-Stuff12K}}} & \myim{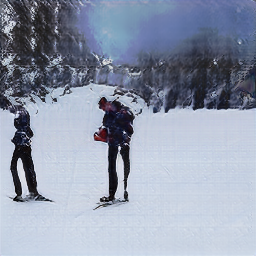} &
    \myim{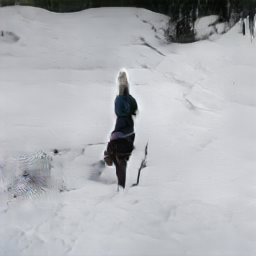} &
    \myim{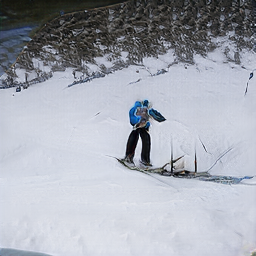}\\
    &\myim{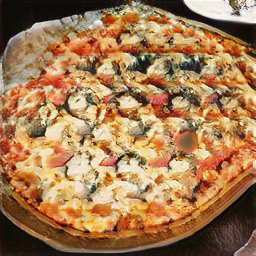} &
    \myim{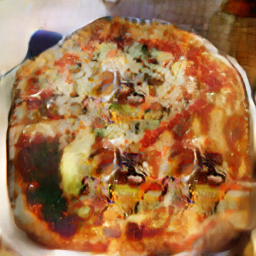} &
    \myim{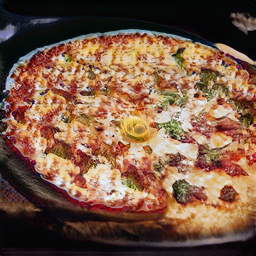} \\
    &\myim{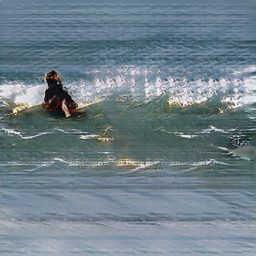} &
    \myim{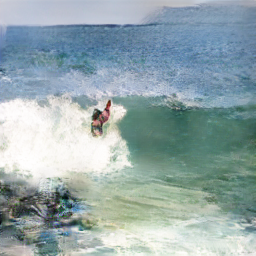} &
    \myim{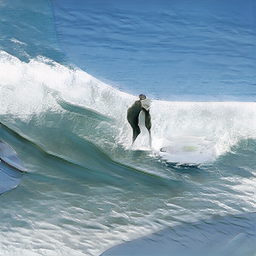} \\
      \end{tabular}
  \end{minipage}
  \hspace{7mm}
  \begin{minipage}[t]{0.62\textwidth}
  \setlength\tabcolsep{0.5 pt}
  \renewcommand{\arraystretch}{0.2}
      \tiny
     \begin{tabular}{cccc}
    & OASIS& SB-GAN& \ours\\
  \myim{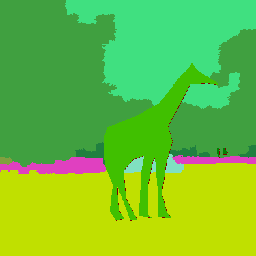} &
    \myim{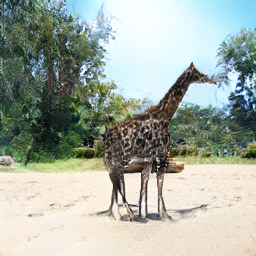}&
    \myim{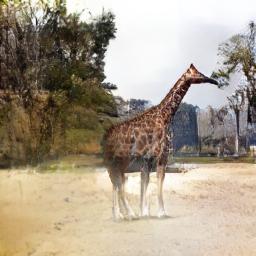}&
    \myim{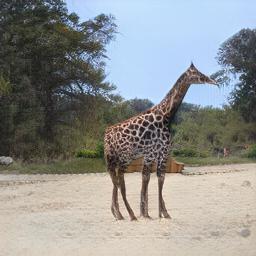}  \\
    \myim{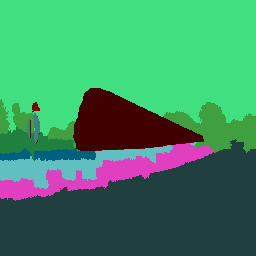} &
    \myim{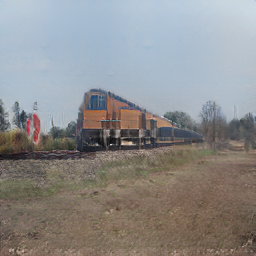}&
    \myim{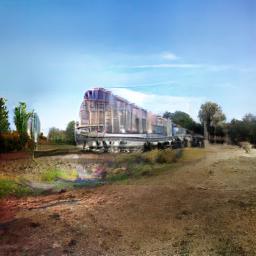}&
    \myim{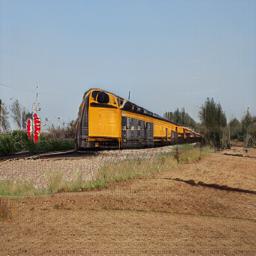}  \\
    \myim{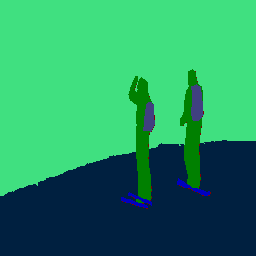} &
    \myim{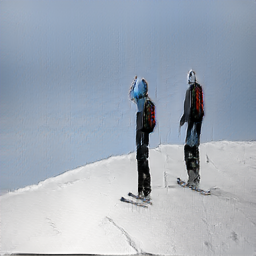}&
    \myim{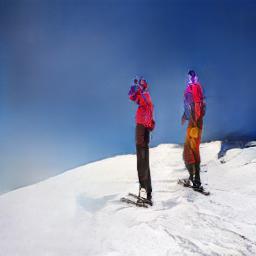}&
    \myim{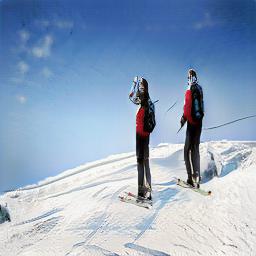}  
     \end{tabular}
\end{minipage}   
\label{fig:adeindoor}
\end{figure}

\subsection{Main results}

\mypar{Limited Data Setting.}
In Table \ref{tab:smalldatamain} we evaluate  \ours in limited data regime in terms of FID, CFID  and mIoU metrics, along with state-of-the-art 
 baselines. On both Cityscapes and ADE-Indoor, SB-GAN  outperforms StyleGAN2  and BigGAN  unconditional models, as well as SPADE on conditional generation, showing that this is a strong baseline  for hybrid generation.

\ours  consistently  improves the hybrid PSP and SB-GAN baselines on all metrics and datasets. 
We observe  poor CFID and mIoU values for PSP on all the datasets, showing that this two-stage approach fails to generate realistic images that adhere to the semantic conditioning. 
\ours also outperforms the unconditional baselines StyleGAN2, StyleMapGAN and BigGAN on all the datasets, with an average improvement of 9.9 points of FID in comparison to the best baseline. 
It shows that the unconditional generation  benefits from the semantic supervision signal  during training. 
When it comes to conditional generation, \ours matches or improves the CFID and mIoU scores of the state-of-the-art semantic synthesis model SPADE and OASIS. 
CFID is improved by 6.1 points on Cityscapes5K, while we notice an improvement of 2.4 points on ADE-Indoor. 
On ADE4K, SPADE is  slightly better than  \ours in CFID by 0.4, but has a worse mIoU value by 1.8. 
In the case of COCO-Stuff12K dataset, \ours is on par with OASIS in CFID, but improves  mIoU by 2.7.

In Figure  \ref{fig:adeindoor} we present samples for the ADE-Indoor and COCO-Stuff12K datasets.  
For the unconditional case, we show samples depicting similar scenes to aid comparison.
In these more diverse datasets, \ours is able to produce images of higher quality containing sharper and more precise details, in both conditional and unconditional settings.
See for example the unconditional samples depicting kitchens  and dining rooms  for ADE-Indoor, or  the conditional giraffe sample for COCO-Stuff12K, in particular the legs and torso of the giraffe, as well as the trees in the background in the same image.

\begin{table}[t]
\caption{Results in the  {\bf Partially annotated setting}. In bold: Best result among  unconditional, conditional, and hybrid models, resp.
}
\tabspacetop
\centering
\setlength{\tabcolsep}{1pt} 
{ \scriptsize
 \begin{tabular}{lrrr|rrr|rrr}
\toprule
   & \multicolumn{3}{c|}{\textbf{Cityscapes}}& \multicolumn{3}{c|}{\textbf{ADE20K}} & \multicolumn{3}{c}{\textbf{COCO-Stuff}}\\
& 
 $\downarrow$\textbf{FID}  &   $\downarrow$\textbf{CFID} &    $\uparrow$\textbf{mIoU}   &
 $\downarrow$\textbf{FID}  &   $\downarrow$\textbf{CFID} &    $\uparrow$\textbf{mIoU}   &
 $\downarrow$\textbf{FID}  &   $\downarrow$\textbf{CFID} &    $\uparrow$\textbf{mIoU}   \\ \midrule
StyleGAN2~\cite{Karras2019stylegan2} & 49.1 & && 43.1  &   &  &  37.0 &   &   \\
StyleMapGAN~\cite{kim21cvpr} & {\bf48.8} & & & \bf 39.6  &   &   & \bf 33.0  &   &   \\
BigGAN~\cite{brock19iclr}  &  67.8 & & & 81.0 & & & 70.5 &  &    \\
\midrule
SPADE~\cite{park19cvpr1} &    &   83.7 & 52.1  & & \bf 43.4 & 34.3  &  &  \bf 42.0 & \bf 28.6 \\
OASIS~\cite{sushko21iclr} &   & \bf 77.4 &  \bf 63.5  & & 49.9 &  \bf 36.1  &  & 77.2 & 23.5 \\
\midrule
SB-GAN~\cite{azadi2019semantic} & 96.1& 93.1  & 50.2 & 82.6   & 45.5  & 31.0 & 103.5 & 57.6 & 19.2  \\
PSP~\cite{richardson21cvpr} &  \bf 49.1 & 111.8  & 13.8 & 43.1  &  147.2 & 2.8 & \bf 37.0 & 168.1 & 2.4  \\
 \ours & 49.6 & \bf 49.3  & \bf 65.2 & \bf 40.4 & \bf 40.1 & \bf 37.6 & 44.4 & \bf 36.8  & \bf 32.1 \\
\bottomrule
\end{tabular}
}
\tabspacebot
\label{tab:semiannotatedAde20kCOCO}
\end{table}

\begin{figure}[t]
\caption{Semantic  synthesis on Cityscapes25K in the {\bf partially annotated setting}.   
}
\def\myim#1{ \includegraphics[width=0.245\textwidth]{figures/#1}}
     \setlength\tabcolsep{0.5 pt}
  \renewcommand{\arraystretch}{0.2}
    \centering
  \begin{tabular}{ccccc}
  & OASIS& SB-GAN & \ours
      \\
\myim{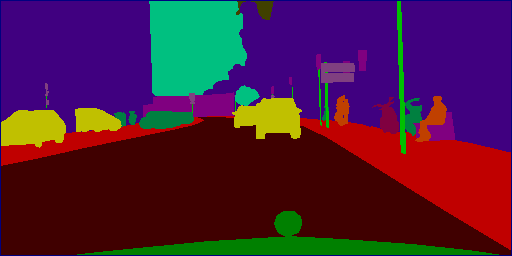} &
\myim{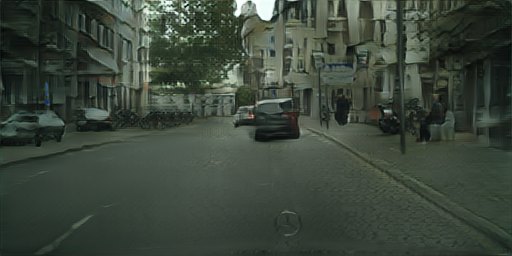} &
\myim{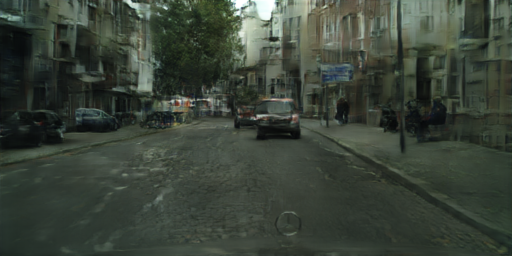} &
\myim{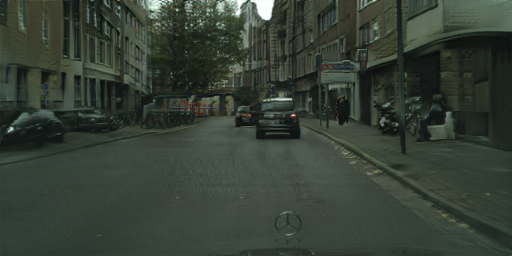}  
\\
\myim{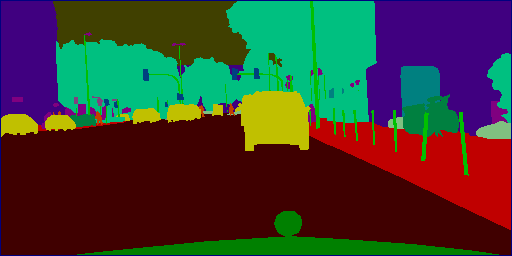} &
\myim{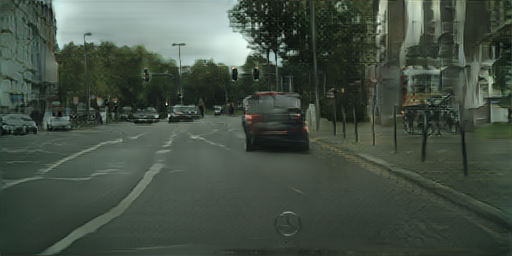} &
\myim{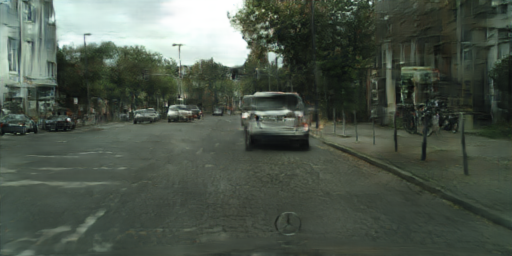} &
\myim{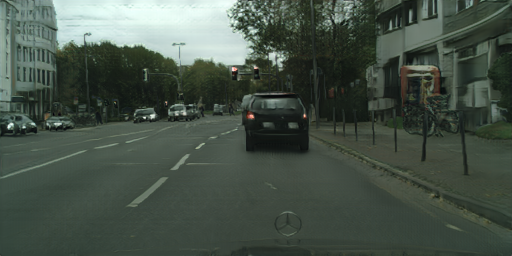}
\\
\myim{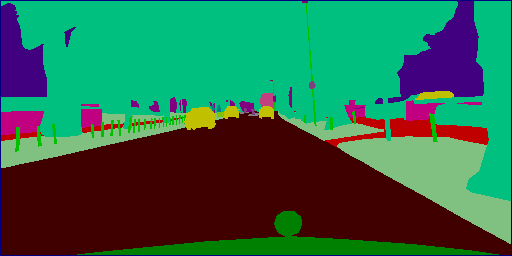} &
\myim{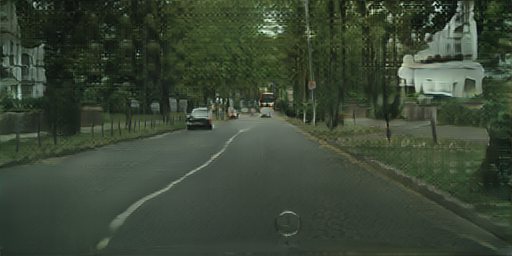} &
\myim{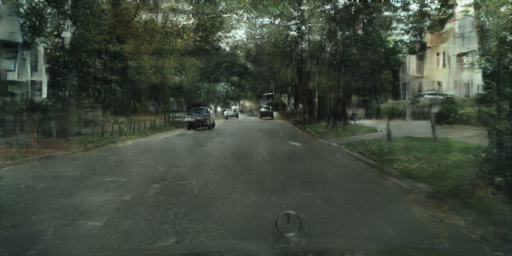} &
\myim{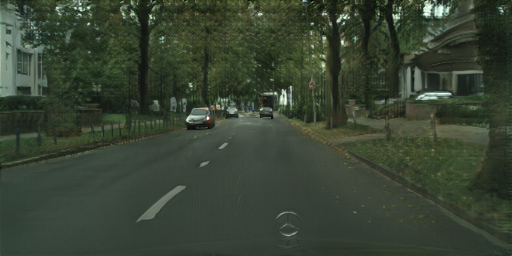}

\end{tabular}
\tabspacebot
\label{fig:cityscapescondsemisup}
\end{figure}

\mypar{Partially annotated Setting.}
We present our results in the partially annotated setting in  Table \ref{tab:semiannotatedAde20kCOCO}.
Also in this setting, OCO-GAN considerably outperforms hybrid baselines in most cases.
Indeed, PSP only works well for unconditional generation, as it is based on a pretrained StyleGAN2 generator (trained on the full unlabeled dataset in this setting), but  performs poorly for semantic  synthesis due to the limited data to learn the style inference network. 
SB-GAN has worse metrics than \ours for both types of generation. While our model is competitive with unconditional baselines in terms of FID, we observe a substantial improvements in conditional generation when comparing to SPADE and OASIS. Indeed, for Cityscapes, ADE20K and COCO-Stuff, \ours is respectively 21.6, 3.3 and 3.0 points of CFID lower, and 9.9, 1.5 and 3.5 points of mIoU higher than the best conditional model. Where  in the limited data setting above  unconditional generation benefits from joint training,  in the current setting  conditional generation improves by jointly training our hybrid architecture. In Figure \ref{fig:cityscapescondsemisup}, we show conditional samples of Cityscapes. We observe sharper details and better quality images for \ours compared to OASIS and SB-GAN, especially when looking at cars and vegetation. 

 \begin{table}[t]
 \caption{Results on the {\bf full datasets}.   Italic: Results  taken from original papers.
On Cityscapes, SB-GAN uses a segmentation net to train with additional segmentations. In bold: Best result among  unconditional, conditional, and hybrid models, resp.
}
\tabspacetop
\centering
\setlength{\tabcolsep}{1pt} 
\resizebox{\textwidth}{!}
{
\scriptsize
 \begin{tabular}{lrrr|rrr|rrr|c}
\toprule
   & \multicolumn{3}{c|}{\textbf{Cityscapes}}& \multicolumn{3}{c|}{\textbf{ADE20K}} & \multicolumn{3}{c|}{\textbf{COCO-Stuff}} & \textbf{Inf.\ time secs.} \\
& 
 $\downarrow$\textbf{FID}  &   $\downarrow$\textbf{CFID} &    $\uparrow$\textbf{mIoU}   &
 $\downarrow$\textbf{FID}  &   $\downarrow$\textbf{CFID} &    $\uparrow$\textbf{mIoU}   &
 $\downarrow$\textbf{FID}  &   $\downarrow$\textbf{CFID} &    $\uparrow$\textbf{mIoU}   &
 $\downarrow$\textbf{Unc./Cond.}   \\ \midrule
StyleGAN2~\cite{Karras2019stylegan2} & 49.1 & && 43.1  &   &  &  37.0 &   & &-/0.02  \\
StyleMapGAN~\cite{kim21cvpr} & {\bf48.8} & & &\bf  39.6  &   &   & \bf 33.0  &   & &-/0.06  \\
BigGAN &  67.8 & & & 81.0 & & & 70.5 &  & & -/0.01   \\
\midrule
SPADE~\cite{park19cvpr1} &    & \it 71.8 & \it 62.3 & & \it 33.9 & \it 38.5 &  & \it 22.6 & \it 37.4 &0.03/-\\
OASIS~\cite{sushko21iclr} &   &\emph{ \textbf{47.7}}&  \textbf{\emph{ 69.3}}  & & \textbf{\emph{28.3}} &  \textbf{\emph{48.8}}  &  &  \textbf{\emph{ 17.0}} & \textbf{\emph{ 44.1}}& 0.01/-\\
\midrule
SB-GAN~\cite{azadi2019semantic}  & \it 63.0 & \it 54.1  & n/a & 62.8  & 35.0 & 34.5 & 89.6 & 42.6 & 19.3 & 0.07/0.03 \\
PSP~\cite{richardson21cvpr} & 49.1 & 99.2 & 15.6 & 43.1 & 146.9 &2.8 & \bf 37.0 & 211.6 & 2.1&  0.02/0.12\\
\ours &  {\bf48.8} &{\bf 41.5} & \bf 69.1& {\bf38.6} &  \bf 30.4 & \bf 47.1 & 38.5  &  \bf  20.9 &  \bf 42.1 & 0.05/0.06  \\
\bottomrule
\end{tabular}
}
\tabspacebot
\label{tab:fullAde20kCOCO}
\end{table}

\begin{figure}[t]
\caption{Unconditional (left) and cond.\ (right) samples from  ADE20K models in  the {\bf full data setting}. 
}
\medskip
\def\myim#1{ \includegraphics[width=16.5mm,height=16.5mm]{figures/#1}}
   \begin{minipage}[t]{0.37\textwidth}
   \setlength\tabcolsep{0.5 pt}
   \renewcommand{\arraystretch}{0.2}
    \tiny
     \begin{tabular}{cccc}
     StyleMapGAN & SB-GAN& \ours\\
     \myim{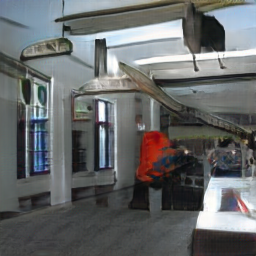} &
\myim{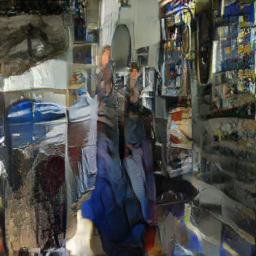} &
\myim{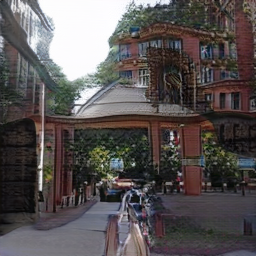}\\
\myim{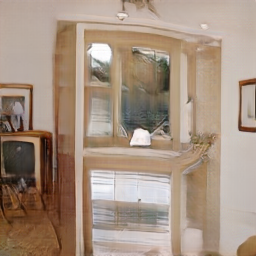} &
\myim{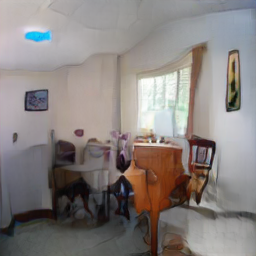} &
\myim{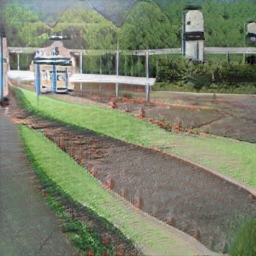}\\
\myim{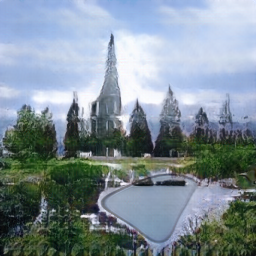} &
\myim{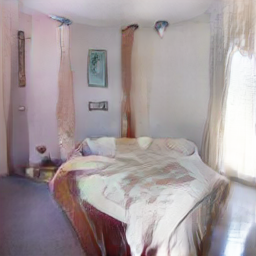} &
\myim{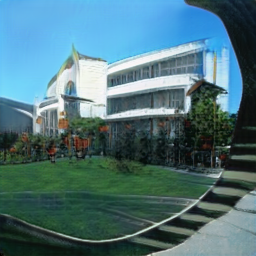}
    
      \end{tabular}
   \end{minipage}
   \hspace{7mm}
   \begin{minipage}[t]{0.62\textwidth}
   \setlength\tabcolsep{0.5 pt}
   \renewcommand{\arraystretch}{0.2}
       \tiny
     \begin{tabular}{cccc}
    & OASIS& SB-GAN& \ours\\
   \myim{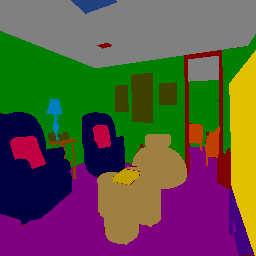} &
    \myim{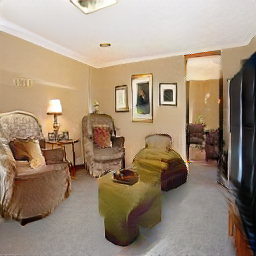}&
    \myim{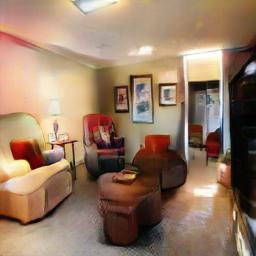}&
    \myim{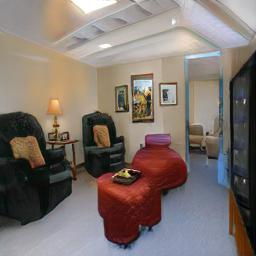}  \\
    \myim{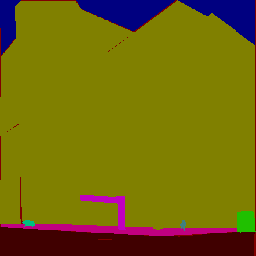} &
    \myim{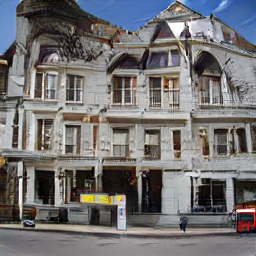}&
    \myim{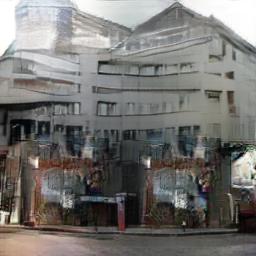}&
    \myim{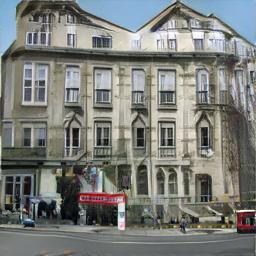}  \\

    \myim{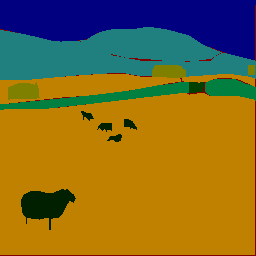} &
    \myim{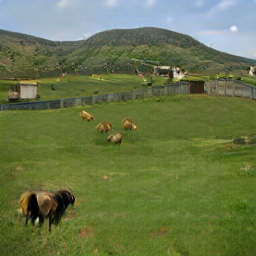}&
    \myim{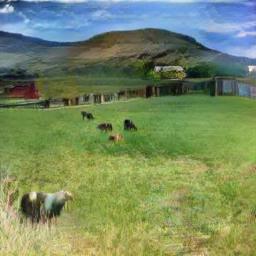}&
    \myim{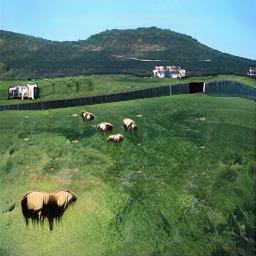}  \\
     \end{tabular}
\end{minipage}   
\tabspacebot
\label{fig:fullade}
\end{figure}

\mypar{Full Data Setting.} We provide results on the full datasets  in Table \ref{tab:fullAde20kCOCO}. 
As in the  limited data and partially annotated experiments, \ours outperforms  the hybrid baselines, SB-GAN and PSP in all cases, except on COCO-Stuff where PSP has slightly better FID.
Although in this setting there is less benefit of addressing both tasks with a single unified model, our results are still competitive with specialized unconditional and conditional generation models. 
For the unconditional task, \ours achieves the best FID values on Cityscapes and ADE20K, but obtains a  higher FID  than StyleGAN2 on COCO-Stuff. 
For  conditional generation, \ours obtains slightly worse scores than OASIS (except for CFID on Cityscapes, where \ours is the best), but outperforms SPADE. 

We present unconditional and conditional synthesized samples of models trained on ADE20k in Figure~\ref{fig:fullade}. We notice the improvement of image quality from SB-GAN to \ours. Indeed, the difference is particularly striking when looking at the building in the second row for semantic image synthesis, which is better rendered by \ours. 
We additionally report inference time and comment on the results in section C.7 of Supplementary Material.

\subsection{Ablation studies}

\mypar{Impact of hybrid training and architecture.}
In \tab{archi}, we conduct an ablation on Cityscapes5K in the limited data regime and Cityscapes25K in the partially annotated setting to separate out the effect of the \ours architecture and that of the joint training of the unconditional and conditional branches. 

For unconditional generation, we see the benefit of hybrid training in the unconditional generation results of \ours in  the \emph{limited data regime}.
We obtain 82.0 FID by training \ours for unconditional generation only, which improves to 57.4 FID when jointly training both (un)conditional branches. 
In the partially annotated setting we do not observe this dramatic improvement, as in this case many more unlabeled images are available during training.

\begin{table}[t]
\caption{Ablation of architectures and training modes. We evaluate models in the limited data  setting with Cityscapes5K and partially annotated setting with Cityscapes25K (with 500 annotated images).
Italics: results   from original paper. 
}
\tabspacetop
\centering
\setlength{\tabcolsep}{1pt} 
{\scriptsize
 \begin{tabular}{lccc|ccc}
 
\toprule
  &  \multicolumn{3}{c|}{\textbf{Cityscapes5K}} &  \multicolumn{3}{c}{\textbf{Cityscapes25K}} \\
 & 
$\downarrow$\textbf{FID}  &   $\downarrow$\textbf{CFID} &    $\uparrow$\textbf{mIoU} &    
$\downarrow$\textbf{FID}  &   $\downarrow$\textbf{CFID} &    $\uparrow$\textbf{mIoU}   \\ \midrule
 \textbf{Unconditional training} \\ 
 \midrule
StyleGAN2  &  79.6  &   &    &   49.1  &   &   \\
StyleMapGAN &   70.4 &   &  &  \bf 48.8  &   &  \\
\ours{ }- uncond.\  only  &  82.0  &   &  &  50.5  &   &   \\
\midrule
\textbf{Conditional training} \\ \midrule
OASIS &   &  \emph 47.7   & \emph 69.3  &   &  77.4   &   63.5 \\
 \ours{ }- cond.\ only &   &  43.7  & \bf 70.3  &   &  76.8  & 57.7  \\
\midrule
\textbf{Hybrid training} \\ \midrule
\ours & {\bf57.4 } & {\bf41.6 } & 69.4  & 49.8  & \bf 55.8 & \bf 65.2   \\
\bottomrule
\end{tabular}
}
\tabspacebot
\label{tab:archi}
\end{table}

For conditional generation, when we only train the conditional branch, \ours improves over OASIS on both datasets and metrics, in particular for the partially annotated setting (CityScapes25K)  where there are only 500 labeled images. Hybrid training  improves  conditional generation in all cases, except in mIoU for Cityscapes5K, where the mIoU degrades by 0.9 points, from 70.3 to 69.4,  to the level of OASIS (69.3). In the partially annotated case the CFID is very strongly by 19.3 points from 76.8 to 57.5 CFID, demonstrating that \ours  effectively leverages  the many unlabeled images available in this setting.  

Moreover, we ran two additional  experiments on Cityscapes5K. In the first, we start training the unconditional part only,  then fix it and train the conditional branch. In the second, we proceed in the reverse order.
In the first case we get a FID of 73.8 and CFID of 126.4 while in the second case we have FID  95.3 and CFID  50.2, which is to compare with a FID  57.4 and CFID  41.6 in the joint training. These results show the 
benefit of our joint training approach, as compared to stage-wise training, as done \eg in PSP~\cite{richardson21cvpr}.

\mypar{Comparison with Adaptive Data Augmentation.}
Data augmentation can boost the performance of generative models, especially when dealing with small datasets,
see \eg~\cite{karras20nips,tran20arxiv,zhao20nips,zhao20arxiv}.
To assess to what extent such methods are complementary to our hybrid approach,  
we perform experiments with Adaptive Data Augmentation (ADA)~\cite{karras20nips} in the limited data setting.  
The results in \tab{augment} show that both StyleGAN2 and \ours benefit from ADA for all datasets.
We observe an improvement of 9.4, 10.2 and 13.4 points  FID when adding ADA to \ours on Cityscapes5K, ADE-Indoor and ADE4K respectively, while for the larger COCO-Stuff-12K the improvement is limited to 1.5 points.

\mypar{Ablation of regularisation terms to train our models.}
We show in Table 5 of Supplementary Material the effects of regularization during training. 

\begin{table}[t]
\caption{
Effect of Adaptive Data Augmentation (ADA) on FID for  StyleGAN2 and \ours (trained in hybrid mode) for the datasets of the limited data setting.
}
\tabspacetop
\centering
\setlength{\tabcolsep}{1pt} 
 \scriptsize
 \begin{tabular}{lrrrr}
\toprule
&  \textbf{City5K} & \textbf{ADE-Ind} & \textbf{ADE4K}  & \textbf{CCS12K}\\ 
\midrule
StyleGAN2    & 79.6  &  87.3  & 95.7  & 77.4  \\
same w/ ADA & 60.6  &   63.9  & 77.8  & \bf48.9  \\
\midrule
\ours       & 57.4  &  69.0  & 80.6  & 51.5  \\
same w/ ADA &{\bf48.0}  & \bf 58.8 &\bf67.2&50.0\\
\bottomrule
\end{tabular}
\tabspacebot
\label{tab:augment}

 \end{table}

\section{Conclusion}
\label{sec:conc}

We propose \ours, a model capable of synthesizing images  with and without conditioning on semantic maps. 
We combine a shared style-based synthesis network,  containing most of the network parameters, with conditional and unconditional style generating networks. 
We  train the model using a U-Net discriminator with a full-image real-fake classification branch, and a per-pixel branch that classifies pixels across semantic classes and an additional ``fake'' label. 

We validate our model on the  Cityscapes, ADE20K, and COCO-Stuff datasets, using three settings: a limited data setting, a semi-supervised setting, and the full datasets.
In all settings our approach outperforms existing hybrid models that can generate both with and without spatial semantic guidance,  and is competitive  with specialized conditional or unconditional state-of-the-art baselines.
In limited data regime our method improves over state-of-the-art unconditional models, by leveraging the additional training signal provided by the labeled images. In the semi-supervised setting we improve over state-of-the-art conditional  models, by using the large pool of unlabeled images to regularize the training.

\clearpage
\bibliographystyle{splncs04}
\bibliography{references2,references}

\clearpage

\appendix

In this supplementary material, we first provide additional details regarding the source codes and datasets used in our experiments (Sec.~\ref{app:assets}). Then, in Sec.~\ref{sec:archi}, we detail the proposed architecture. In Sec.~\ref{sec:implem}, we provide and evaluate some implementation details. We report additional qualitative (Sec.~\ref{sec:qualitative}) and quantitative  (Sec.~\ref{sec:quantitative}) results.

\section{Assets and licensing information}
\label{app:assets}

In Table~\ref{table:assets}, we provide the links to the  datasets and code repositories used in this work, and list their licenses in Table~\ref{table:licenses}.

\begin{table*}
\centering
\caption{Links to the assets used in the paper.}
\label{table:assets} 

\scriptsize
\begin{tabular}{ll}
\toprule
\bf Datasets\\
COCO-Stuff~\cite{caesar18cvpr} & \url{https://cocodataset.org/} \\ 
Cityscapes~\cite{cordts16cvpr} & \url{https://www.cityscapes-dataset.com/} \\ 
Ade20K~\cite{zhou17cvpr}& \url{http://groups.csail.mit.edu/vision/datasets/ADE20K/}\\
\midrule
\bf Code repositories\\
BigGAN~\cite{brock19iclr} & \url{https://github.com/ajbrock/BigGAN-PyTorch} \\
StyleGAN2~\cite{Karras2019stylegan2} & \url{https://github.com/rosinality/stylegan2-pytorch} \\
StyleGAN2 with ADA~\cite{karras20nips} & \url{https://github.com/NVlabs/stylegan2-ada-pytorch}\\
StyleMapGAN ~\cite{kim2021stylemapgan} & \url{https://github.com/naver-ai/StyleMapGAN}\\
PSP~\cite{richardson21cvpr} & \url{https://github.com/eladrich/pixel2style2pixel} \\
SB-GAN~\cite{azadi2019semantic} &  \url{https://github.com/azadis/SB-GAN} \\
SPADE~\cite{park19cvpr1} &  \url{https://github.com/NVlabs/SPADE} \\
OASIS~\cite{sushko21iclr} &  \url{ https://github.com/boschresearch/OASIS} \\
\bottomrule
\end{tabular}

\end{table*}

\begin{table}
\centering
\caption{Assets licensing information.}
\label{table:licenses}
\setlength{\tabcolsep}{1pt} 

\scriptsize
\begin{tabular}{ll}
\toprule
\bf Datasets\\
COCO-Stuff~\cite{caesar18cvpr} & \url{https://www.flickr.com/creativecommons} \\ 
Cityscapes~\cite{cordts16cvpr} & \url{https://www.cityscapes-dataset.com/license} \\ 
Ade20K~\cite{zhou17cvpr}& MIT License\\
\midrule
\bf Code repositories\\
BigGAN~\cite{brock19iclr} & MIT License \\
StyleGAN2~\cite{Karras2019stylegan2} & \url{https://github.com/rosinality/stylegan2-pytorch/blob/master/LICENSE} \\
StyleGAN2 with ADA ~\cite{karras20nips}  & NVIDIA Source Code License \\
StyleMapGAN ~\cite{kim2021stylemapgan}& \url{https://github.com/naver-ai/StyleMapGAN/blob/main/LICENSE}\\
PSP~\cite{richardson21cvpr} &  MIT License\\
SB-GAN~\cite{azadi2019semantic} & \url{https://github.com/azadis/SB-GAN/blob/master/LICENSE.md}  \\
SPADE~\cite{park19cvpr1} &  \url{https://github.com/NVlabs/SPADE/blob/master/LICENSE.md} \\
OASIS~\cite{sushko21iclr} &  \url{https://github.com/boschresearch/OASIS/blob/master/LICENSE} \\
\bottomrule
\end{tabular}

\end{table}

\section{Architecture details}
\label{sec:archi}

\begin{figure}
\caption{Generator and discriminator network architectures.}
\centering
\includegraphics[width=.6\linewidth]{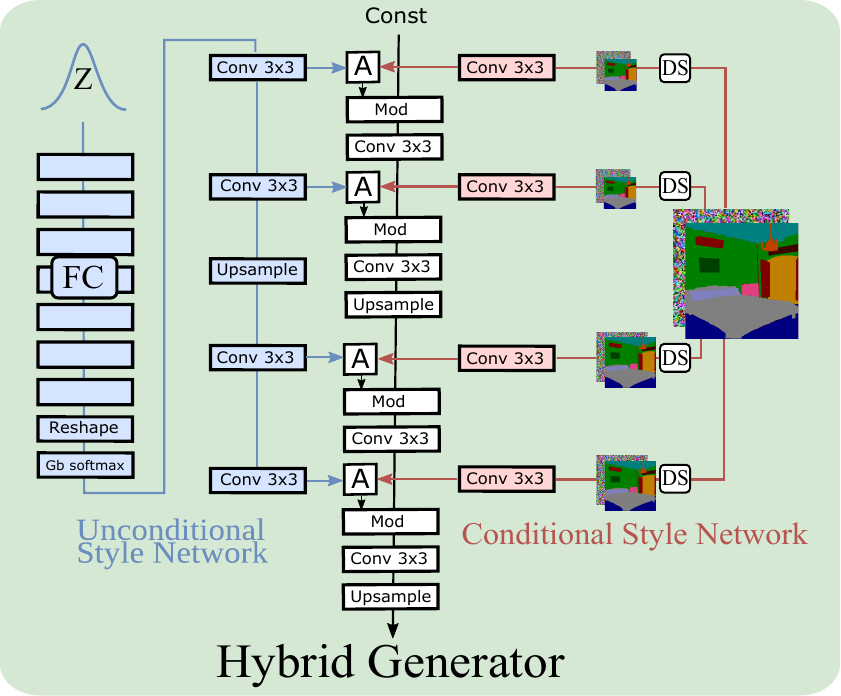}
\\
\vspace{1mm}
\includegraphics[width=.6\linewidth]{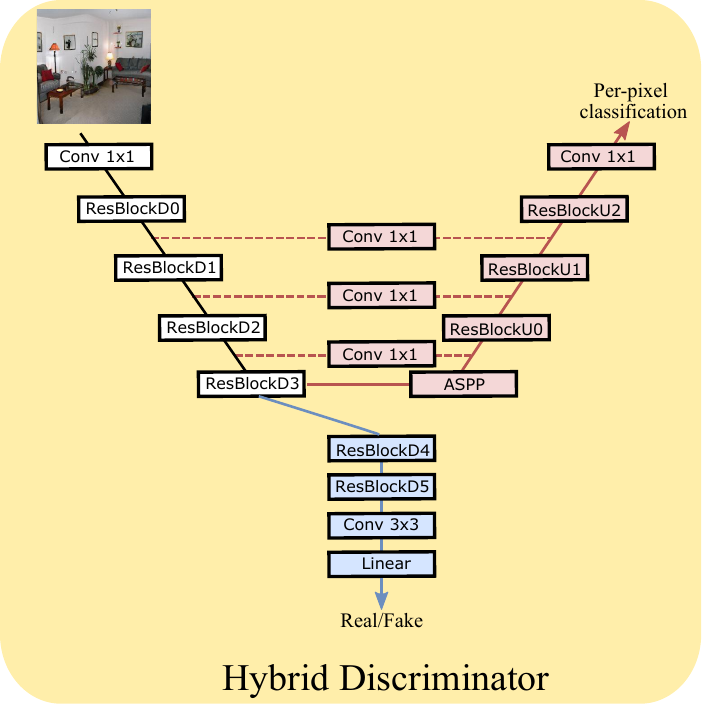}

\label{fig:archis}
\end{figure}

In Fig.~\ref{fig:archis}, we show a detailed representation of our generator and discriminator architectures. 

In the generator (top panel), our synthesis network (in white, middle) takes style inputs at different resolutions either from the  unconditional (in blue, left) or conditional (in red, right) style network. 
These styles are mapped to modulation parameters $\gamma$ and $\beta$ through the module ``A'', a learned affine transform. 
Then, feature maps are modulated in the ``Mod'' block. 
We first normalize them and then apply an element-wise multiplication and addition using $\gamma$ and $\beta$. 
At each spatial resolution of the feature maps, there are two blocks of modulation and convolution layers followed by a upsampling operation. 
In parallel we have a skip connection that by-passes the two blocks at each resolution, not represented in the figure for clarity. 
It consists of one convolution (without modulation) and an upsampling operation. 
The output of the latter is added to the activation maps of the main branch containing modulated convolutions after the upsampling block.

Our discriminator (bottom panel) contains residual blocks indicated by ``ResBlockD'' in the common backbone (in white). 
Then, our unconditional head (in blue) consists of two other residual blocks followed by a convolution and linear layers to predict whether the image is fake or real. 
Our conditional head (in red) first transforms the output of the common encoder with an {\it atrou} spatial pyramid pooling (ASPP)~\cite{DBLP:journals/corr/ChenPSA17}. It contains {\it atrou} convolutions at different rates operating in parallel, whose outputs are concatenated and upsampled. We then have residual blocks indicated by ``ResBlockU'' blocks which progressively upsample the feature maps. The latter also takes as input lateral connections from the encoder through skip connections shown with dashed lines. 

\section{Implementation details}
For all the models, we monitor FID during training. We end the training either when the FID starts diverging or when it stops improving. The hyperparameters of \ours and the baselines are selected  based on the final FID value at the end of training, and we select the configuration that leads to the best FID.

\label{sec:implem}
\subsection{\ours and StyleGAN2 hyperparameters}
For training \ours and StyleGAN2 baselines, we adopt the default hyperparameters of StyleGAN2, using a R1 regularization in the encoder of the discriminator with $\gamma=10$ and an exponential moving average of the generator with value 0.999. For optimization, we take an Adam optimizer~\cite{kingma15iclr} with learning rate 0.002, $b_1 = 0.0$ and $b_2 = 0.99$. 
We use a batch size of $\BS= 64$ for training StyleGAN2 on all the datasets, except ADE20K where we
obtain better FID with $\BS = 32$. 
For \ours, we have a mixed batch of conditional and unconditional samples. In the limited and full data settings, the conditional batch size is 16 for all the datasets, except COCO-Stuff where we use a batch size of 32. 
In the semi-supervised setting, as there is a smaller amount of annotated samples, we found empirically better results when reducing the conditional batch to 4 for all the datasets, otherwise the conditional branch tends to overfit and converge faster than the unconditional one. The unconditional batch size is 32. 
We also find it beneficial to apply a weight of 2 on the unconditional loss when training \ours with ADE20K and COCO-Stuff datasets.
Regarding data augmentation, we apply random horizontal flip, similar to OASIS, SPADE and StyleGAN2, except when we specify that we use Adaptive Data Augmentation (ADA)~\cite{karras20nips}. 
Our code is based on the unofficial repository of StyleGAN2 cited in Table \ref{table:assets}. 
We evaluate StyleGAN2 based on this repository, while we train StyleGAN2 with ADA using the official repository from NVIDIA.

\subsection{Hyperparameters of the other baselines}

For SPADE and OASIS, we follow their experimental settings, using the same training hyperparameters of the full dataset when using subsets of it.

For BigGAN, we explored batch size $\BS \in \{64,128, 256\}$, channel multiplier $\ch \in \{32,48\}$ and learning rate $ \lr \in \{ 4\mathrm{e}{-4}, 1\mathrm{e}{-4}, 4\mathrm{e}{-5}\}$. We used $\ch = 48$, $\lr = 1\mathrm{e}{-4} $ and $\BS= 256$ for ADE20K and COCO-Stuff. For ADE-Indoor, ADE4K, Cityscapes5K, Cityscapes25K and COCO-Stuff12K, we used $\BS=128$. For Cityscapes25K, ADE-Indoor and COCO-Stuff12k we adopted  $\ch = 32$, $\lr=1\mathrm{e}{-4} $, while we have $\ch = 32$, $\lr = 4\mathrm{e}{-4} $ for ADE-Indoor and $\ch = 48$, $\lr = 1\mathrm{e}{-4} $ for Cityscapes5K. 

The PSP model is trained on top of a pretrained StyleGAN2 generator obtained using the hyperparameters described in the previous section.
Note that Richardson \etal~\cite{richardson21cvpr} initialize the PSP backbone network with a pretrained model on face recognition. Our preliminary experiments show better performance with random initialization on our datasets that include complex scenes not related to facial images. Therefore, we employ random initialization in all our experiments with PSP baseline. 
For semi-supervised experiments, we use a pretrained StyleGAN2 trained on all the dataset, and we train PSP encoder on the available labeled data.
We use the same  loss weighting as Richardson \etal  to train PSP on ADE20K, COCO-Stuff and Cityscapes, \ie a $\ell_2$ loss with weight 1, a LPIPS loss with weight 0.8, and a regularization loss on the latent vectors with weight 0.005. 
The optimizer is a Ranger optimizer with learning rate 0.0001. 

In \cite{azadi2019semantic} SBGAN was evaluated on Cityscapes5K, Cityscapes25K with additional segmentations and ADE-Indoor. We additionally train it on ADE20K, COCO-Stuff, COCOStuff12K and ADE4K, training the SPADE generator with the same experimental setup as described in \cite{park19cvpr1}.  For the segmentation map synthesis network and end-to-end training parts we use the same batch size, learning rate and number of epochs as for ADE-Indoor dataset. When running semi-supervised experiments, we train the semantic synthesizer and SPADE generator with the limited amount of annotated labels and train the unconditional discriminator with all the dataset during the finetuning phase.

\begin{table}
\caption{Impact of Gumbel Softmax (GS) measured on ADE4K.}
\small
\centering
\setlength{\tabcolsep}{3pt} 
\begin{tabular}{lccc}
\toprule
 & $\downarrow$\textbf{FID}  &   $\downarrow$\textbf{CFID} &    $\uparrow$\textbf{mIoU}   \\ \midrule
\ours w/o GS   & 125 $\pm$ 0.7&  44.4 $\pm$ 0.0 & 31.0 $\pm$ 0.0 \\
\ours w/ GS & {\bf80.6 $\pm$ 0.6} &  {\bf43.8 $\pm$ 0.0} & {\bf36.1 $\pm$ 0.0}\\
\bottomrule
\end{tabular}

\label{table:gbsoftmax}
\end{table}

\subsection{Additional experiments on CelebA-HQ}

We complete our experimental evaluation in the three settings with experiments on CelebA-HQ dataset \cite{celeba}, which contains facial images of celebrities with their annotated masks having 19 different classes. We use the standard training/validation splits, which consists of 24,183 annotated training images and 2993 validation data. We compare \ours with the specialized unconditional baseline Stylegan2 and the specialized conditional model OASIS as well as the hybrid model PSP. For PSP, we take their pretrained network for segmentation-to-image trained on CelebA-HQ dataset available in github to compute metrics. Otherwise we trained from scratch the other benchmarks to compare with \ours.
In the limited data regime, we train models using 3k conditional and unconditional images, while in the partially annotated case, we use the full unconditional dataset, ie. 24k images, with 500 annotated data (choosing a similar amount of data as what is used in Table 1 and 2 for the other datasets in the main paper).

\begin{table}[t]
 \caption{Results on CelebA-HQ in the 3 different settings: limited, partially annotated, full data regime. In bold: Best result among  unconditional, conditional, and hybrid models, resp.}
\centering
\setlength{\tabcolsep}{1pt} 
{
\scriptsize
 \begin{tabular}{lrrr|rrr|rrr}
\toprule
 \textbf{CelebA-HQ}  & \multicolumn{3}{c|}{\textbf{Full dataset}}& \multicolumn{3}{c|}{\textbf{Part. annot}} & \multicolumn{3}{c}{\textbf{Limited dataset}}\\
& 
 $\downarrow$\textbf{FID}  &   $\downarrow$\textbf{CFID} &    $\uparrow$\textbf{mIoU}   &
 $\downarrow$\textbf{FID}  &   $\downarrow$\textbf{CFID} &    $\uparrow$\textbf{mIoU}   &
 $\downarrow$\textbf{FID}  &   $\downarrow$\textbf{CFID} &    $\uparrow$\textbf{mIoU}   \\ \midrule
StyleGAN2~\cite{Karras2019stylegan2} &16.4& && 16.4 &   &  & 33.1 &   &   \\

\midrule

OASIS~\cite{sushko21iclr} &   &14.8& \bf 76.9  & & 62.9 & 68.2 &  & 29.8&  72.1\\
\midrule
PSP~\cite{richardson21cvpr} & 16.4 & 50.3 & 68.4 & 16.4 & 80.4 & 61.7 & 33.1 & 70.1 & 60.4 \\
\ours & \bf 13.4   & \bf 14.4  & \bf 76.9& \bf 14.8& \bf 30.9  & \bf 71.1 & \bf 30.2  & \bf 22.9  & \bf 73.4\\
\bottomrule
\end{tabular}
}

\label{tab:celebA}
\end{table}
The results of these experiments are reported in Table~\ref{tab:celebA}.
In the three different settings, we obtain better results in terms of mIoU, FID, CFID compared to the hybrid baseline PSP. We show a significant gain in CFID in the partially annotated setting with respect to the specialized conditional baseline OASIS, which shows that our hybrid training pipeline is beneficial for conditional image generation. In the limited data setting, \ours surpasses all specialized baselines, ie. OASIS and StyleGAN2 on FID, CFID and mIoU.

\subsection{Ablation on regularization terms used during our training pipeline}

We conduct an ablation in Table~\ref{tab:abl_modif} to study the impact of each regularization term we use during training our model. We show that we obtain a loss of performance in FID, CFID and mIoU when removing R1 and LabelMix losses as well as spectral normalization in the encoder of the discriminator. 

 \begin{table}[t]
\caption{
Ablation of regularisation terms used to train our models.
}

\centering
\setlength{\tabcolsep}{1pt} 
 \scriptsize
 \begin{tabular}{lrrr}
\toprule
\textbf{Ablation on Cityscapes5K} &  $\downarrow$\textbf{FID}  &   $\downarrow$\textbf{CFID} &   $\uparrow$\textbf{mIoU} \\ 
\midrule
\ours  &  \bf 57.4  & \bf 41.6 & \bf 69.4  \\
\ours w/o R1 loss &  170.8 & 47.6 & 67.6\\
\ours w/o LabelMix loss   &  58.2  &  43.6 & 67.8 \\
\ours w/ Spectral Norm in Enc.   &  71.1  & 50.5& 68.8\\
\bottomrule
\end{tabular}
\label{tab:abl_modif}
 \end{table}

\subsection{Evaluation}
We evaluate our models with Fr\'{e}chet Inception Distance (FID)~\cite{heusel17nips} for image quality and diversity on conditional and unconditional synthesized images. This score is computed by using the same pretrained model as OASIS~\cite{sushko21iclr}, an Inception v3 network. We extract features from this network on a reference set containing real images and a synthesized one containing fake images from the generator. 
In our setting, we use the same reference set for conditional and unconditional evaluation, which is the validation split, containing respectively 500, 433, 2000 and 5000 images for Cityscapes, ADE-Indoor, ADE20K and COCO-Stuff. 
To evaluate models trained on random subsets of these datasets, we use the same validation set.

In the semantic conditional image synthesis, we additionally evaluate models with mean  Intersection-over-Union (mIoU), measuring consistency between generated images and their conditioning semantic maps. Like in OASIS, we calculate this metric using a pretrained semantic segmentation network, which is UperNet101~\cite{zhou2018semantic} for ADE20K and its subsets, DeepLabV2 ~\cite{DBLP:journals/corr/ChenPKMY14} for COCO-Stuff, and multiscale DRN-D-105~\cite{Yu_2017_CVPR} for Cityscapes dataset.

\subsection{Impact of Gumbel-Softmax.}

We experimentally observed that adding Gumbel-Softmax activations~\cite{jang17iclr,maddison17iclr} on every style map was beneficial when jointly learning unconditional and conditional tasks for all the three settings. 
See Table \ref{table:gbsoftmax} for results of   generation  on ADE4K.

\subsection{Compute resources}

We use eight NVIDIA V100 32GB GPUs to train our \ours  models. 
Because \ours uses a mixed batch of conditional and unconditional samples during training, the total batch is larger than that of StyleGAN2 and OASIS.  
OASIS is trained with 4 GPUs while StyleGAN2 requires 2 GPUs on ADE20K, ADE-Indoor, ADE4K (with batch size 32 at resolution 256$\times$256), 4 GPUs on COCO-Stuff and COCO-Stuff12K
(with batch 64 at resolution 256$\times$256) and 8 GPUs on Cityscapes (with batch 64 at resolution 256$\times$512).

We provide in Table 3 of main paper the running time at inference to generate one image, either unconditionally or conditionally, averaged across 20 images. 
\ours has a similar generation speed as StyleMapGAN, on which we based our architecture. 
Inference in \ours takes longer than in some specialized baseline models, but it is faster SB-GAN for unconditional generation and PSP for conditional one. 
Indeed, SB-GAN uses an encoder module for unconditional generation on top of SPADE, and PSP trains also a conditional module on top of StyleGAN2, which explain the longer inference times.

\section{Additional qualitative results}
\label{sec:qualitative}

Table \ref{tab:figlist} lists figures gathering additional qualitative results for the  datasets studied in our paper. We show additional unconditional and conditional  samples of \ours and compare it to the other competitive baselines. 

We also show diversity of conditional samples when conditioned on the same label map for ADE20K in Figure~\ref{fig:ade20kdiversity}.

\begin{table}
\caption{List of figures.}
\centering
\setlength{\tabcolsep}{1pt} 
\footnotesize
\begin{tabular}{lc}
\toprule
\textbf{Dataset} & \textbf{Figures}  \\
    \midrule
    \bf Limited data setting\\
  
            ADE-Indoor & Figure~\ref{fig:adeindoor_supmat} \\
        ADE4K &  Figure~\ref{fig:ade4k_supmat}\\
         COCO-Stuff12K & Figure~\ref{fig:coco12k_supmat}\\
          \midrule   
          \bf Semi-supervised setting\\
          Cityscapes25K & Figure~\ref{fig:cityscapes25ksemisup} (cond.)\\
        ADE20K  &  Figure~\ref{fig:ade20ksemisup_supmat} (cond.)\\
        COCO-Stuff  &  Figure~\ref{fig:cocosemisup_supmat} (cond.)\\
        \midrule   
        \bf Full datasets\\
         Cityscapes25K & 
        Fig.~\ref{fig:cityscapescondmoredata} (cond.)\\
        ADE20K  & Figure~\ref{fig:ade20k_supmat}\\
        COCO-Stuff  & Figure~\ref{fig:coco_supmat}\\
\bottomrule
\end{tabular}

\label{tab:figlist}
\end{table}

\begin{figure*}
\caption{Randomly selected unconditional (left)  and conditional (right) samples for  ADE-Indoor in limited data setting.  SB-GAN unconditional images are taken from original paper.}

\def\myim#1{ \includegraphics[width=16.5mm,height=16.5mm]{figures/#1}}
 \vspace{0.3cm}
   \begin{minipage}[t]{0.37\textwidth}
   \setlength\tabcolsep{0.5pt}
   \renewcommand{\arraystretch}{0.2}
   \tiny
    \centering
     \begin{tabular}{ccc}
     StyleMapGAN&SB-GAN&\ours\\
\myim{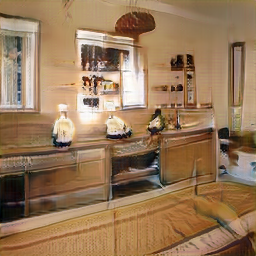} &
\myim{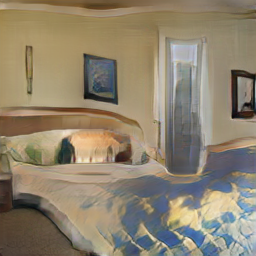} &
\myim{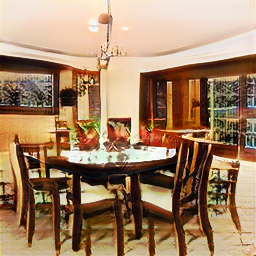}\\
\myim{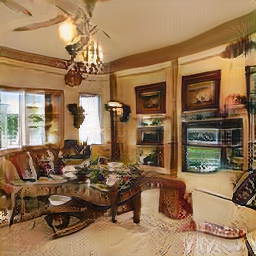} &
\myim{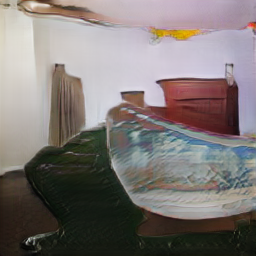} &
\myim{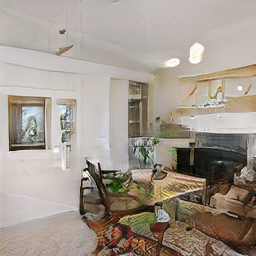}\\
\myim{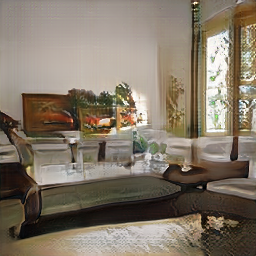} &
\myim{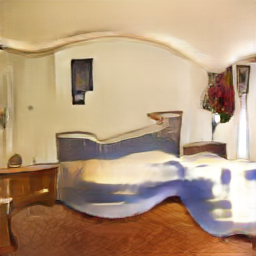} &
\myim{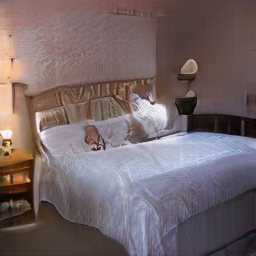}\\
\myim{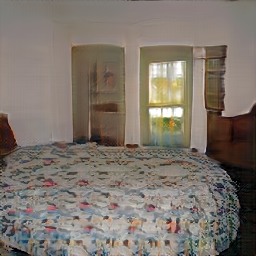} &
\myim{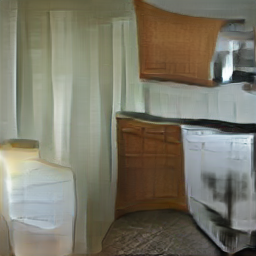} &
\myim{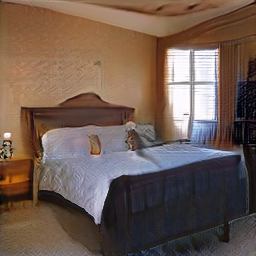}\\
\myim{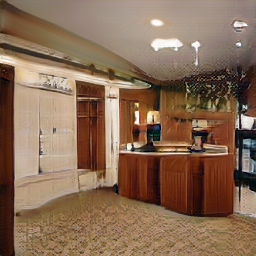} &
\myim{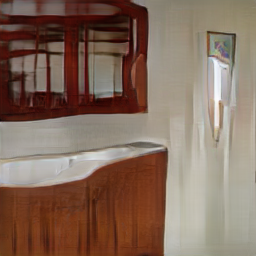} &
\myim{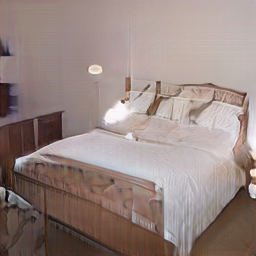}\\

      \end{tabular}
   \end{minipage}
   \hfill
   \begin{minipage}[t]{0.62\textwidth}
     \centering
   \setlength\tabcolsep{0.5 pt}
   \renewcommand{\arraystretch}{0.2}
   \tiny
     \begin{tabular}{cccc}
    & SPADE &SBGAN & \ours\\
\myim{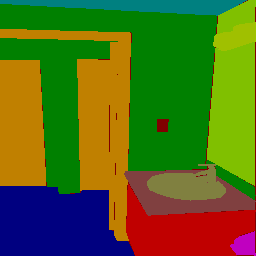} &
\myim{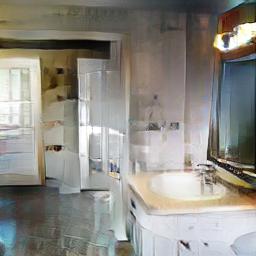} &

\myim{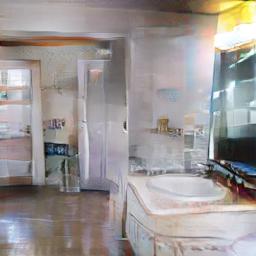}&
\myim{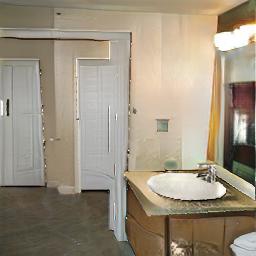}  \\
\myim{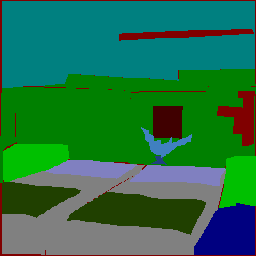} &
\myim{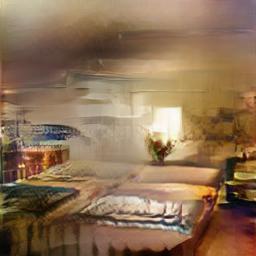} &

\myim{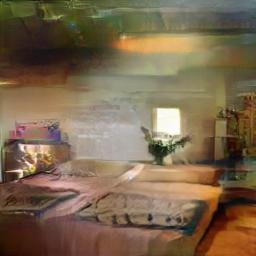}&
\myim{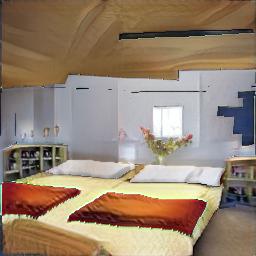}  \\
\myim{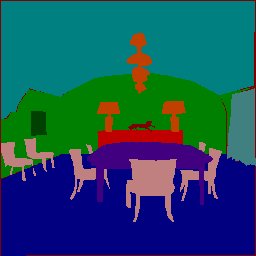} &
\myim{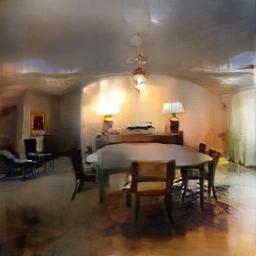} &

\myim{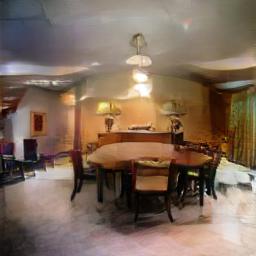}&
\myim{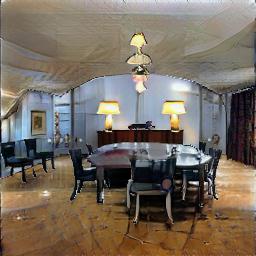}  \\
\myim{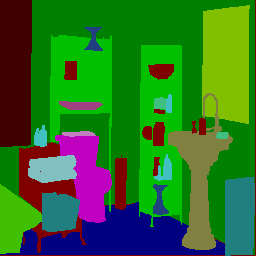} &
\myim{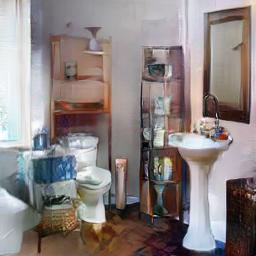} &

\myim{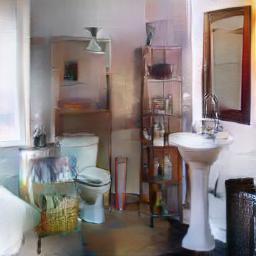}&
\myim{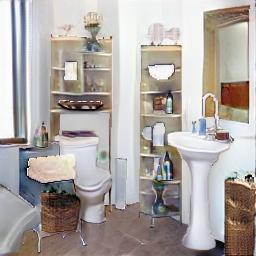}  \\
\myim{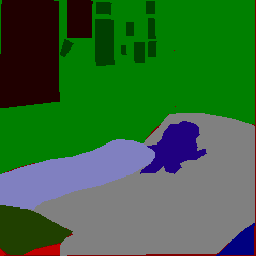} &
\myim{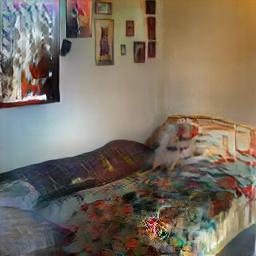} &

\myim{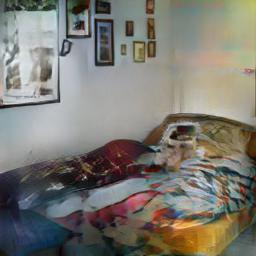}&
\myim{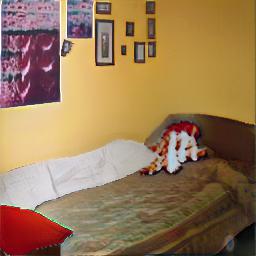}  \\

     \end{tabular}
\end{minipage}    
\label{fig:adeindoor_supmat}
\end{figure*}

\begin{figure*}
\caption{Randomly selected unconditional (left)  and conditional (right) samples for ADE4K in limited data setting. }
 \def\myim#1{ \includegraphics[width=16.5mm,height=16.5mm]{figures/#1}}
 \vspace{0.3cm}
   \begin{minipage}[t]{0.37\textwidth}
   \setlength\tabcolsep{0.5pt}
   \renewcommand{\arraystretch}{0.2}
   \tiny
    \centering
     \begin{tabular}{ccc}
     StyleMapGAN&SB-GAN&\ours\\
\myim{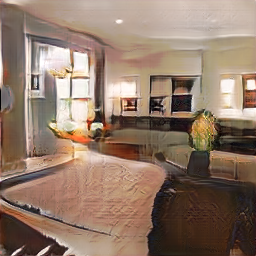} &
\myim{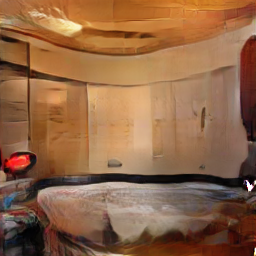} &
\myim{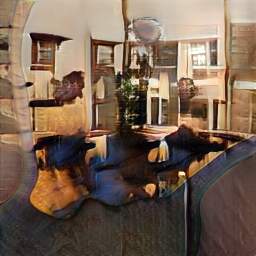}\\
\myim{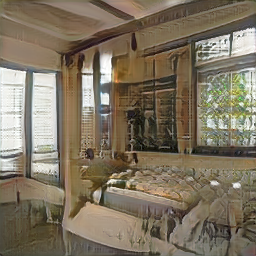} &
\myim{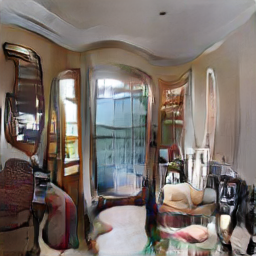} &
\myim{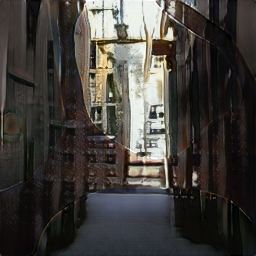}\\
\myim{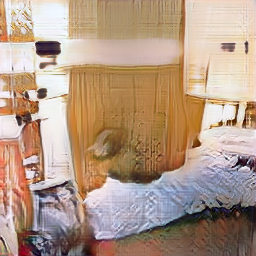} &
\myim{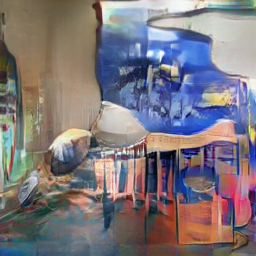} &
\myim{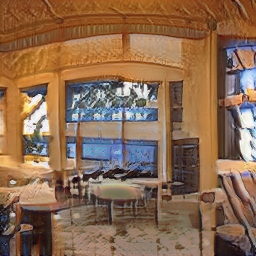}\\
\myim{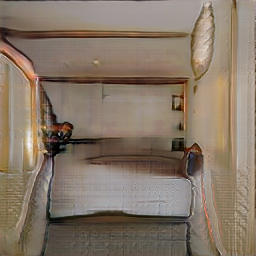} &
\myim{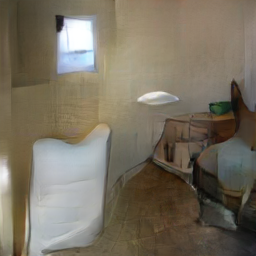} &
\myim{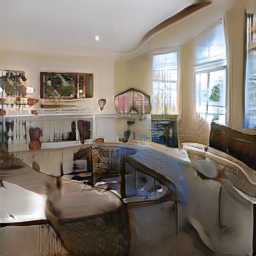}\\
\myim{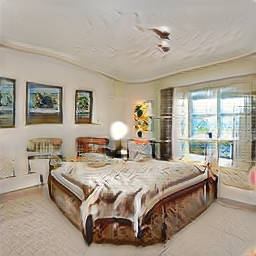} &
\myim{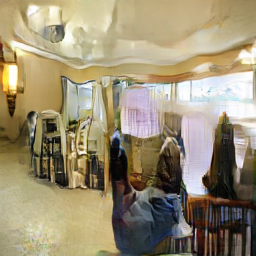} &
\myim{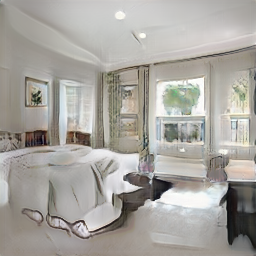}\\

      \end{tabular}
   \end{minipage}
   \hfill
   \begin{minipage}[t]{0.62\textwidth}
     \centering
   \setlength\tabcolsep{0.5 pt}
   \renewcommand{\arraystretch}{0.2}
   \tiny
     \begin{tabular}{ccccc}
    & SPADE & SBGAN & \ours\\
\myim{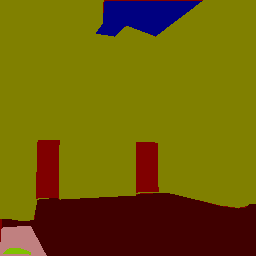} &
\myim{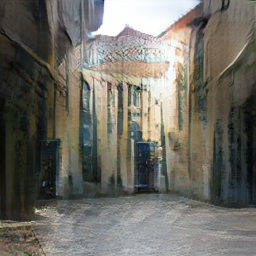} &

\myim{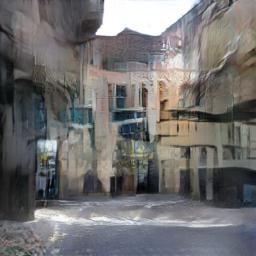} &
\myim{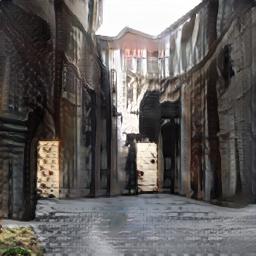}\\
\myim{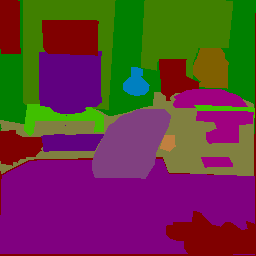} &
\myim{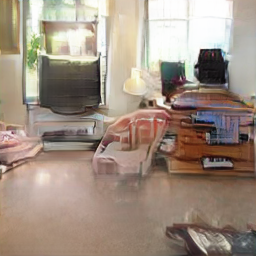} &

\myim{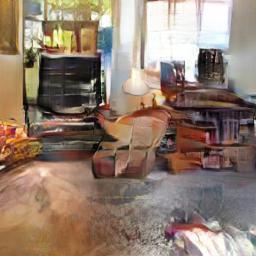} &
\myim{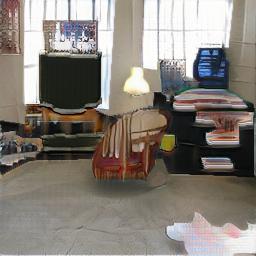}\\
\myim{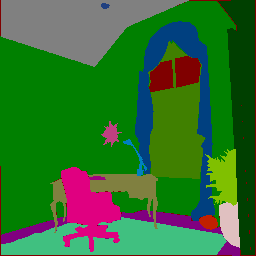} &
\myim{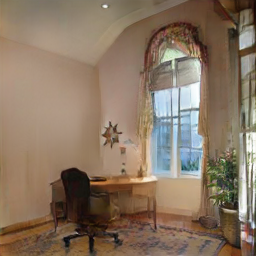} &

\myim{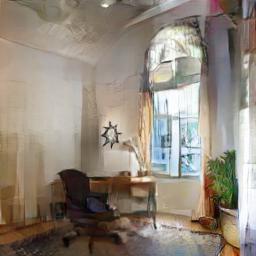} &
\myim{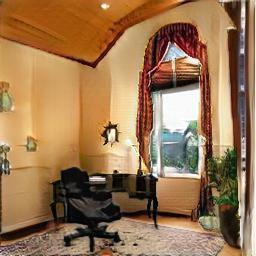}\\
\myim{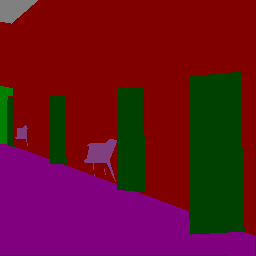} &
\myim{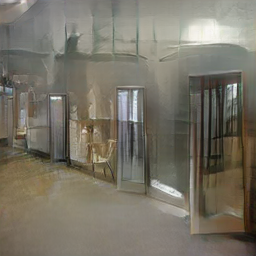} &

\myim{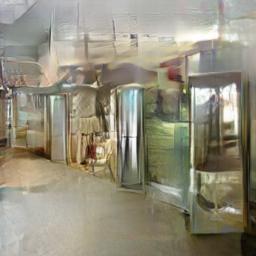} &
\myim{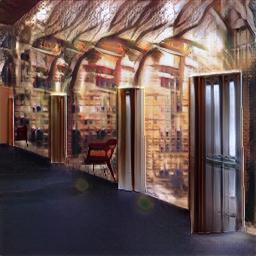}\\
\myim{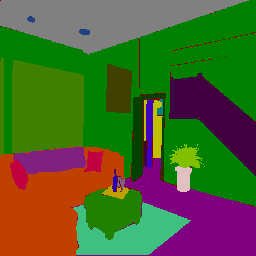} &
\myim{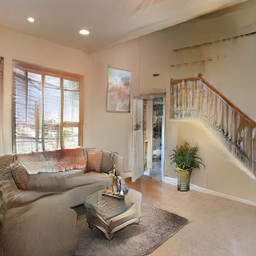} &

\myim{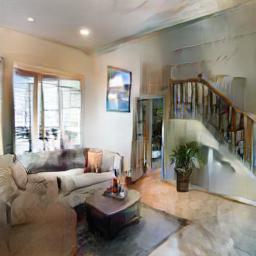} &
\myim{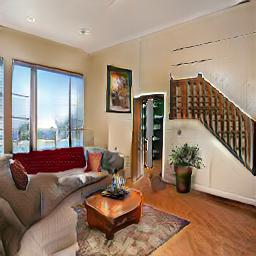}
\\

     \end{tabular}
\end{minipage}    
\label{fig:ade4k_supmat}
\end{figure*}

\begin{figure*}
\caption{Randomly selected unconditional (left)  and conditional (right) samples for  COCO-Stuff12K in limited data setting. }
\def\myim#1{ \includegraphics[width=16.5mm,height=16.5mm]{figures/#1}}
\vspace{0.3cm}
   \begin{minipage}[t]{0.37\textwidth}
   \setlength\tabcolsep{0.5 pt}
   \renewcommand{\arraystretch}{0.2}
   \tiny
    \centering
     \begin{tabular}{cccc}
     StyleMapGAN & SB-GAN& \ours\\
    \myim{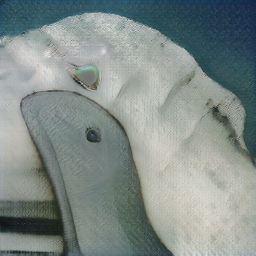} &
    \myim{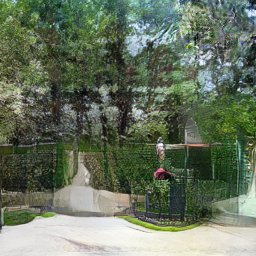} &
    \myim{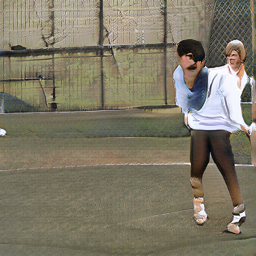}\\
     \myim{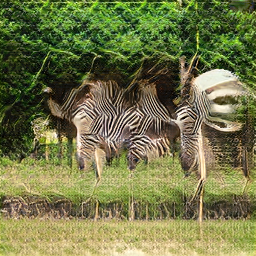} &
    \myim{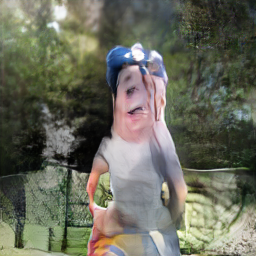} &
    \myim{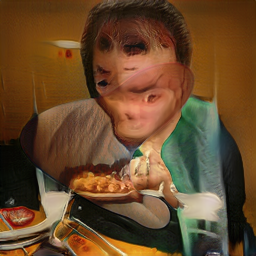}\\
     \myim{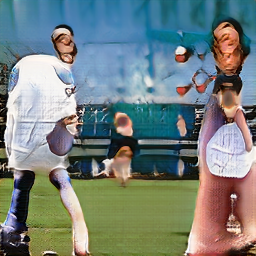} &
    \myim{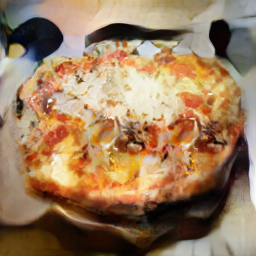} &
    \myim{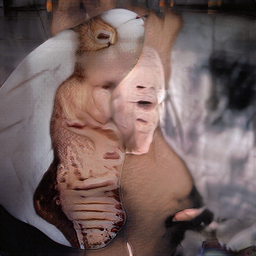}\\
     \myim{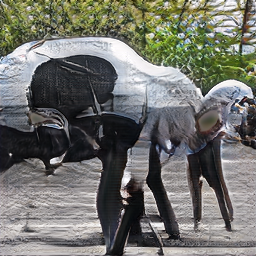} &
    \myim{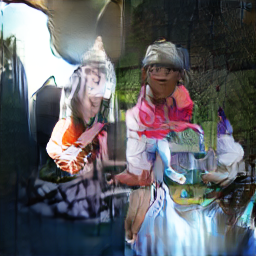} &
    \myim{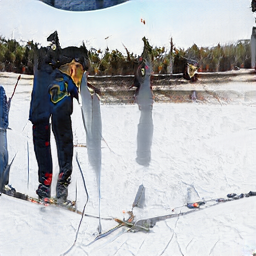}\\
     \myim{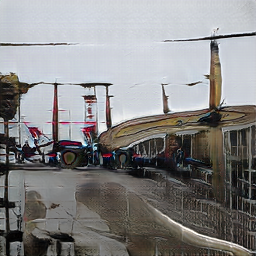} &
    \myim{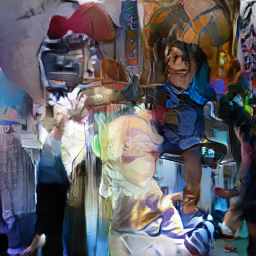} &
    \myim{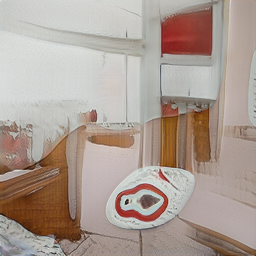}\\
    
      \end{tabular}
   \end{minipage}
   \hfill
   \begin{minipage}[t]{0.62\textwidth}
     \centering
   \setlength\tabcolsep{0.5 pt}
   \renewcommand{\arraystretch}{0.2}
   \tiny
     \begin{tabular}{ccccc}
    & OASIS& SB-GAN& \ours\\
    \myim{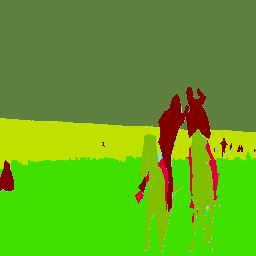}&
    
    \myim{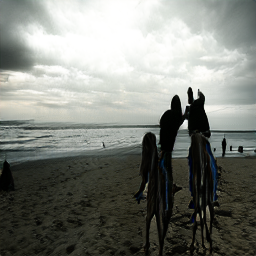}&
    \myim{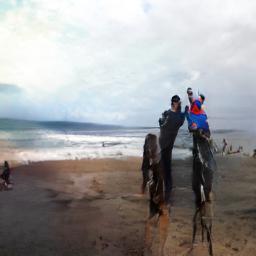}&
    \myim{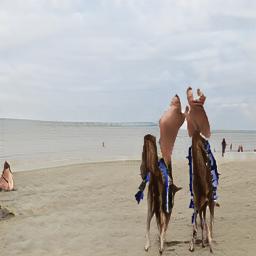}  \\
    \myim{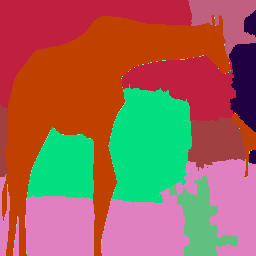}&
    
    \myim{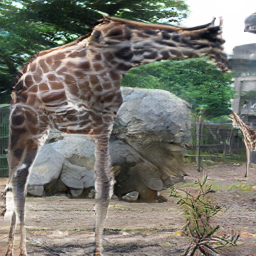}&
    \myim{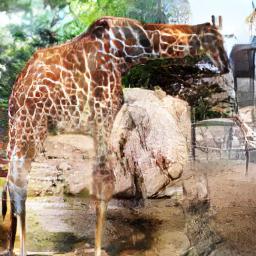}&
    \myim{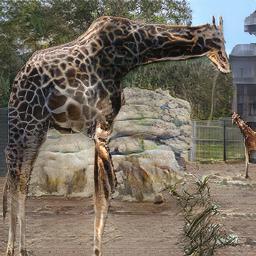}  \\
    \myim{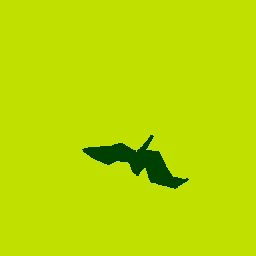}&
    
    \myim{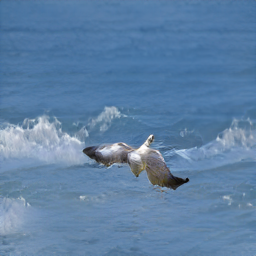}&
    \myim{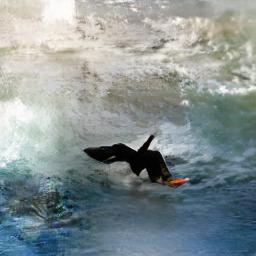}&
    \myim{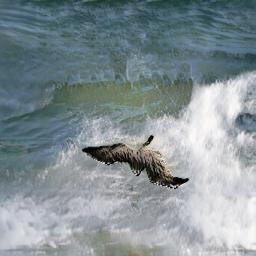}  \\
    \myim{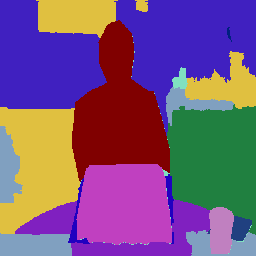}&
    
    \myim{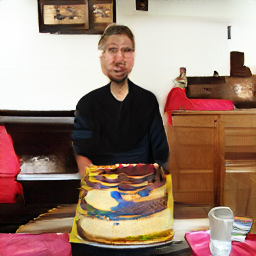}&
    \myim{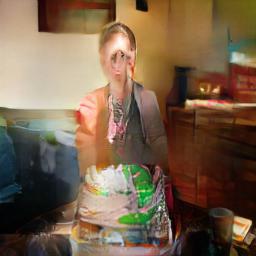}&
    \myim{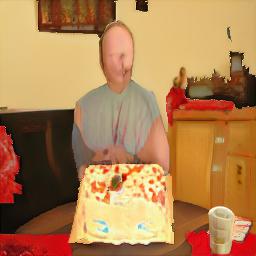}  \\
    \myim{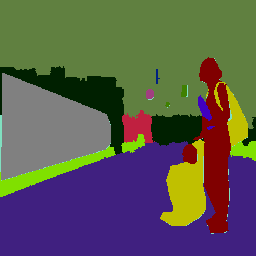}&
    
    \myim{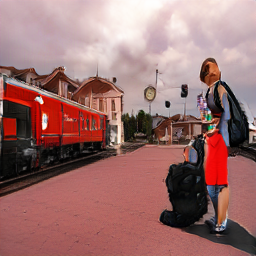}&
    \myim{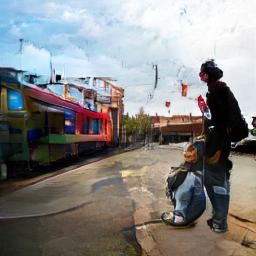}&
    \myim{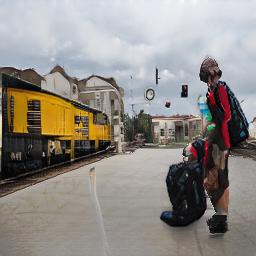}  \\
     \end{tabular}
\end{minipage}    
\label{fig:coco12k_supmat}
\end{figure*}

\begin{figure*}
\caption{Randomly selected samples of semantic image synthesis on Cityscapes25K in semi-supervised setting.}
\def\myim#1{ \includegraphics[width=0.25\textwidth]{figures/#1}}
     \setlength\tabcolsep{0.5 pt}
   \renewcommand{\arraystretch}{0.2}
   \vspace{0.3cm}
    \centering
   \begin{tabular}{ccccc}
   & OASIS& SB-GAN & \ours\\
\myim{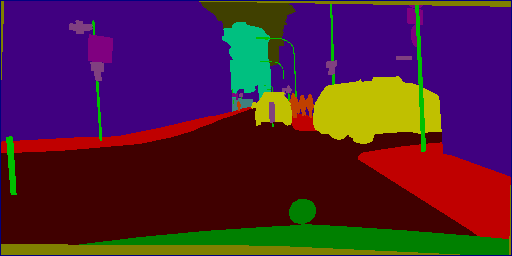} &
\myim{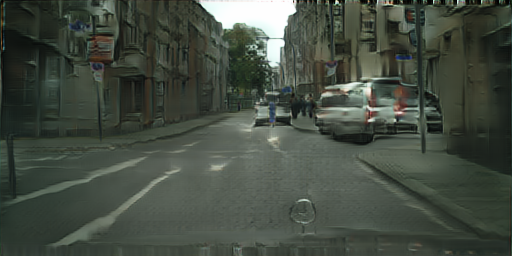} &
\myim{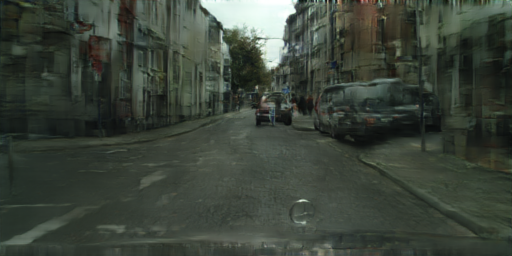} &
\myim{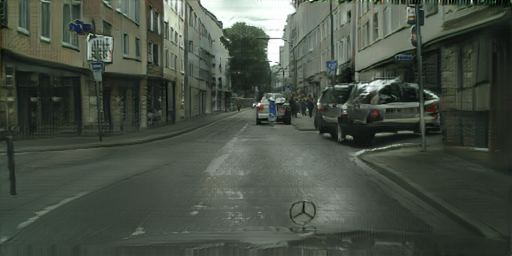} 
    \\
\myim{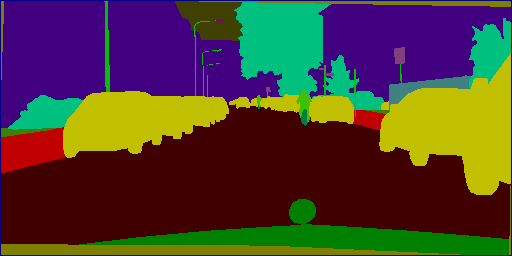} &
\myim{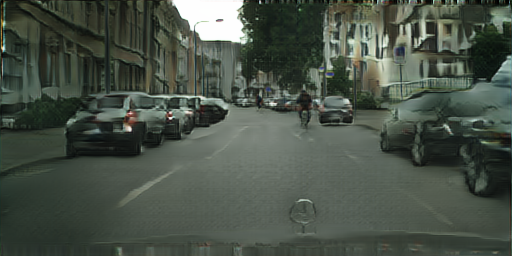} &
\myim{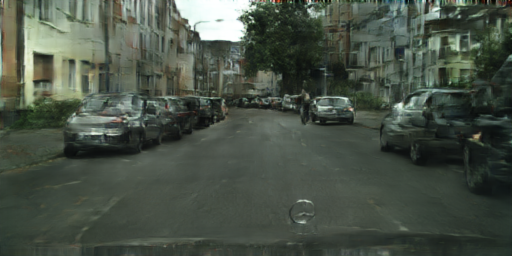} &
\myim{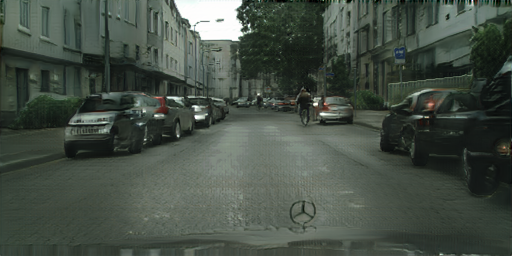} 
      \\
\myim{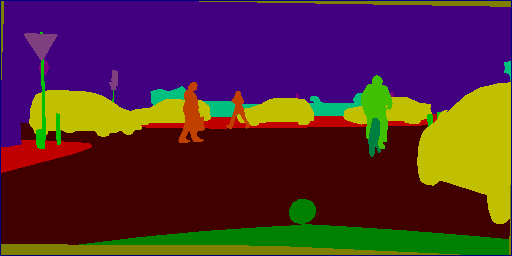} &
\myim{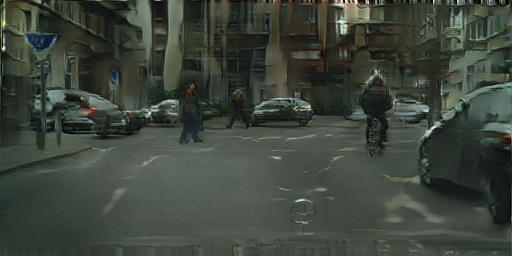} &
\myim{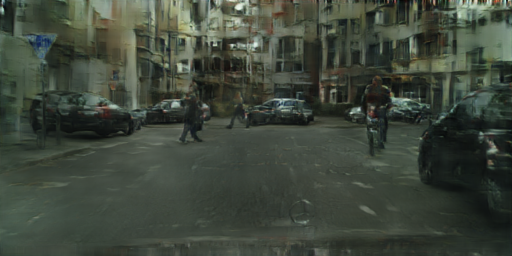} &
\myim{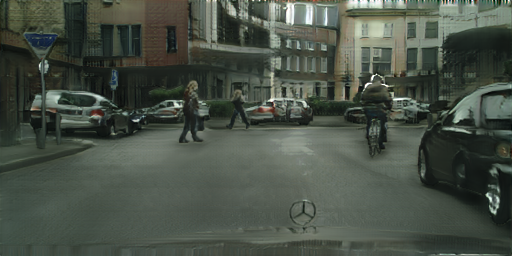} 
\\
\myim{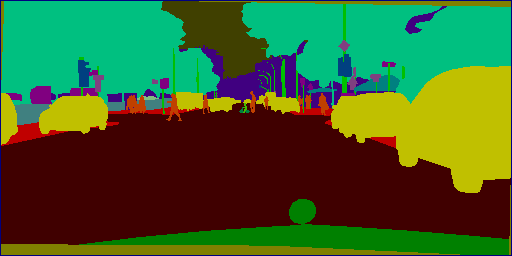} &
\myim{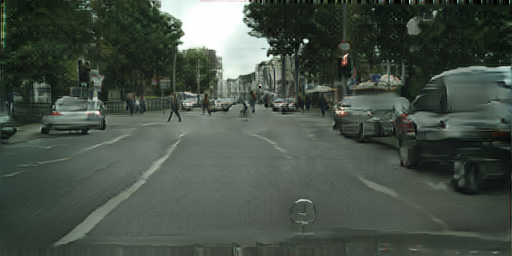} &
\myim{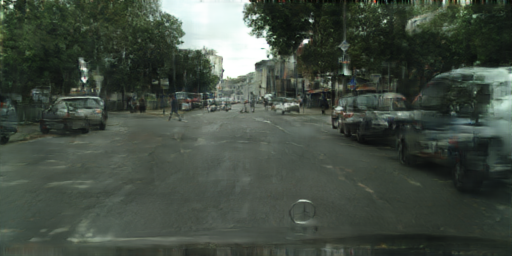} &
\myim{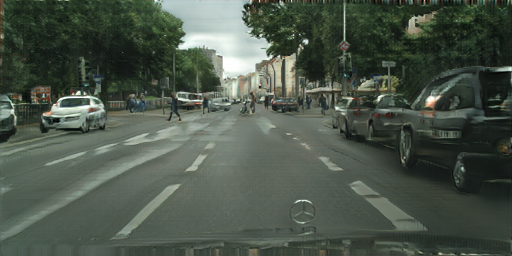} 
\\
\myim{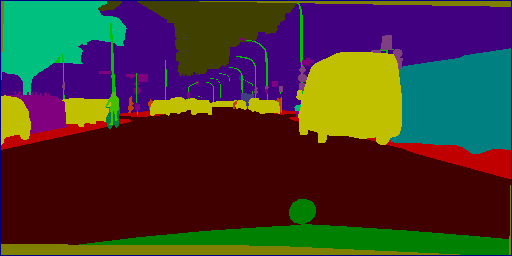} &
\myim{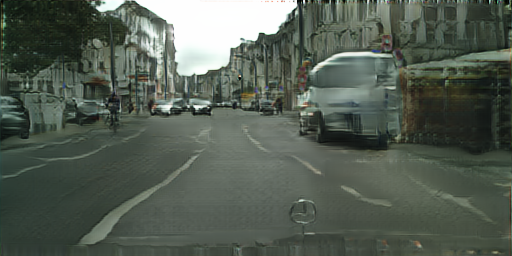} &
\myim{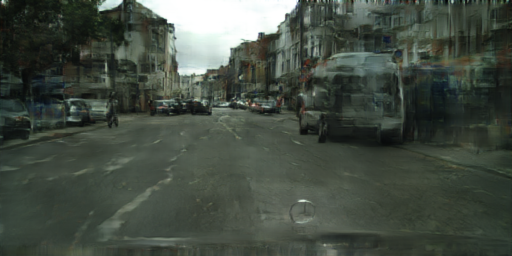} &
\myim{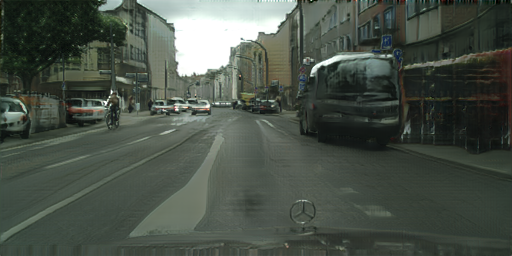} 

\end{tabular}
\label{fig:cityscapes25ksemisup}
\end{figure*}

\begin{figure*}
\caption{Randomly selected conditional  samples for ADE20K in semi-supervised setting. }
 \def\myim#1{ \includegraphics[width=20mm,height=20mm]{figures/#1}}
 \vspace{0.3cm}
     \centering
   \setlength\tabcolsep{0.5 pt}
   \renewcommand{\arraystretch}{0.2}
     \begin{tabular}{ccccc}
    & SPADE & OASIS& SBGAN & \ours\\
\myim{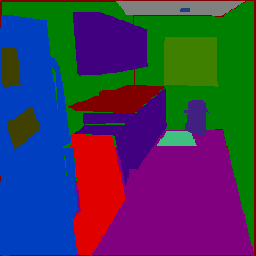} &
\myim{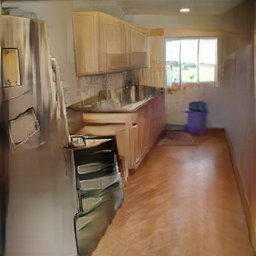} &
\myim{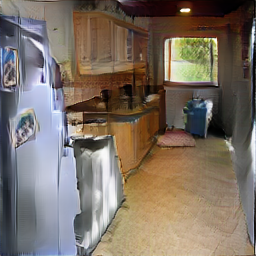}&
\myim{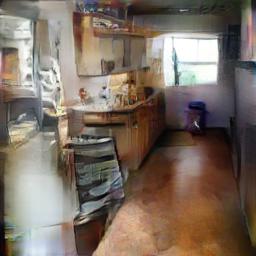} &
\myim{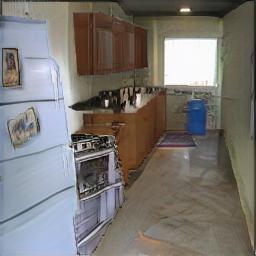}\\
\myim{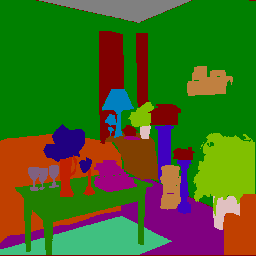} &
\myim{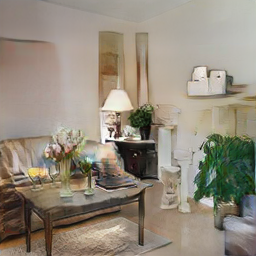} &
\myim{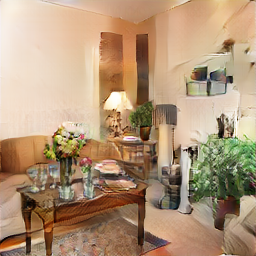}&
\myim{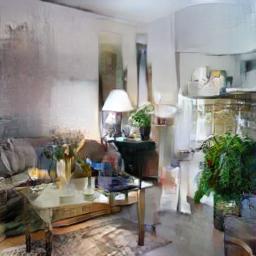} &
\myim{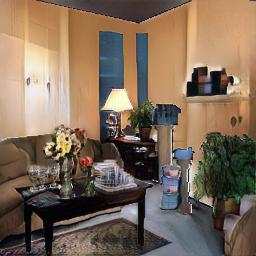}\\
\myim{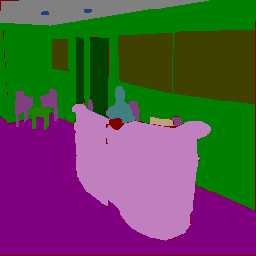} &
\myim{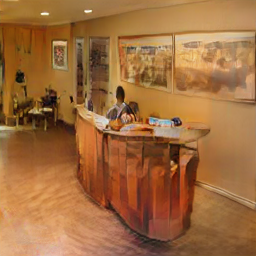} &
\myim{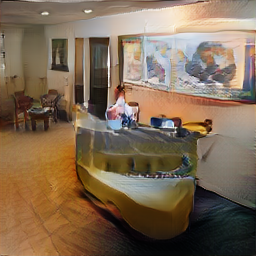}&
\myim{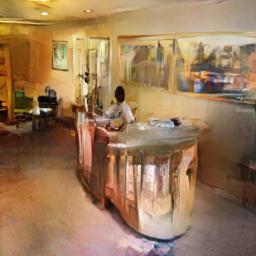} &
\myim{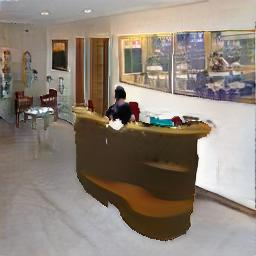}\\
\myim{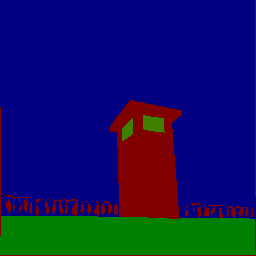} &
\myim{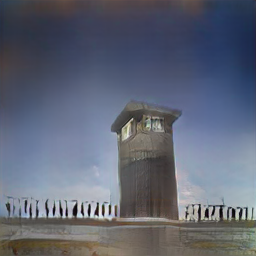} &
\myim{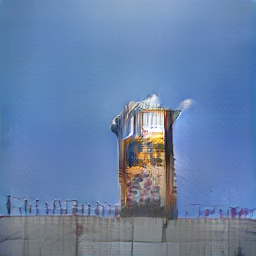}&
\myim{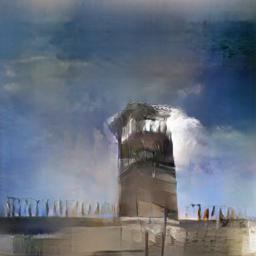} &
\myim{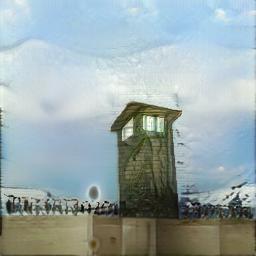}\\
\myim{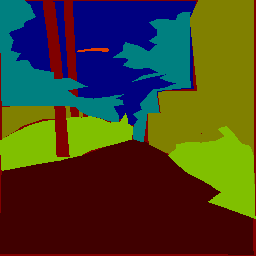} &
\myim{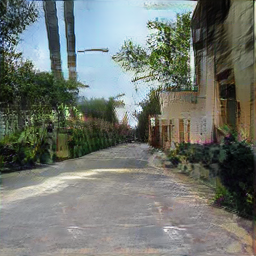} &
\myim{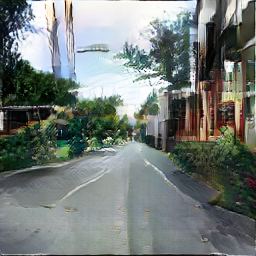} &
\myim{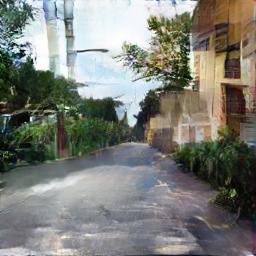}  &
\myim{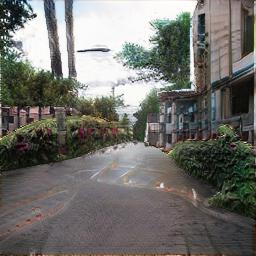} 
     \end{tabular}
\label{fig:ade20ksemisup_supmat}
\end{figure*}

\begin{figure*}
\caption{Randomly selected conditional  samples for COCO-Stuff in semi-supervised setting. }
 \def\myim#1{ \includegraphics[width=20mm,height=20mm]{figures/#1}}
 \vspace{0.3cm}
     \centering
   \setlength\tabcolsep{0.5 pt}
   \renewcommand{\arraystretch}{0.2}
     \begin{tabular}{ccccc}
    & SPADE & OASIS& SBGAN & \ours\\
\myim{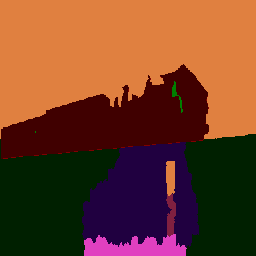} &
\myim{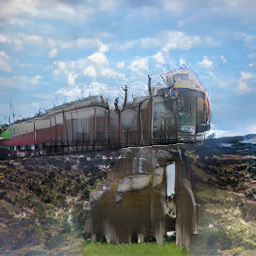} &
\myim{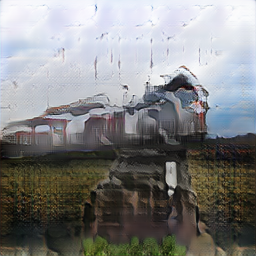}&
\myim{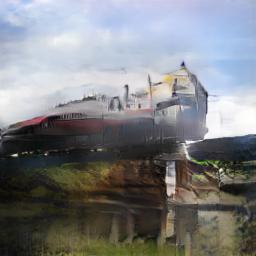} &
\myim{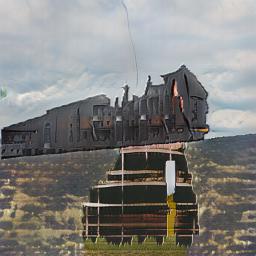}\\
\myim{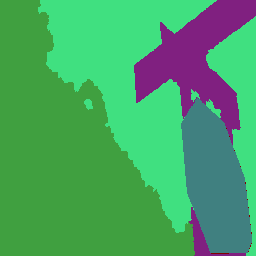} &
\myim{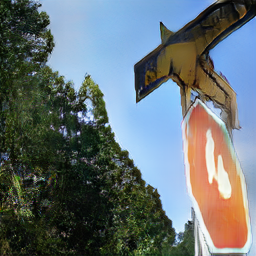} &
\myim{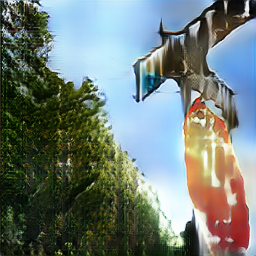}&
\myim{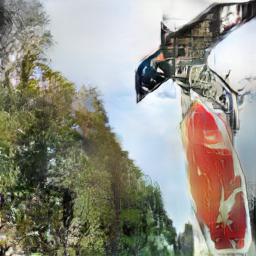} &
\myim{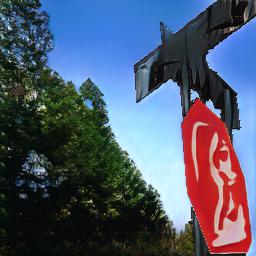}\\
\myim{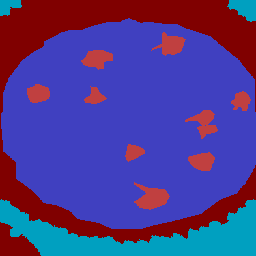} &
\myim{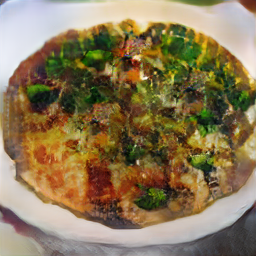} &
\myim{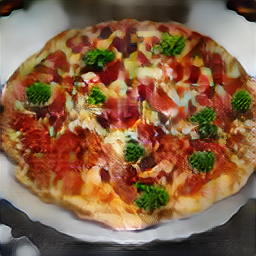}&
\myim{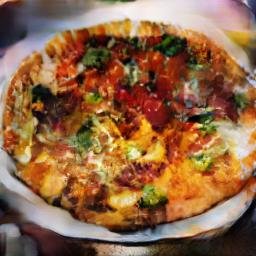} &
\myim{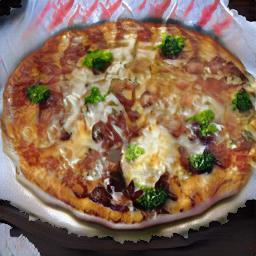}\\
\myim{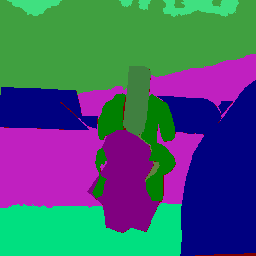} &
\myim{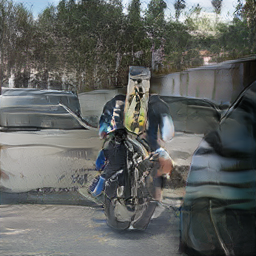} &
\myim{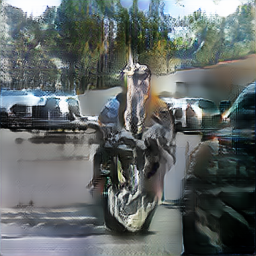}&
\myim{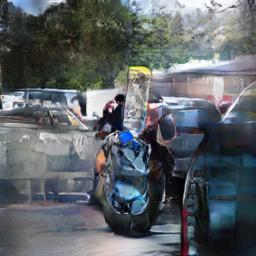} &
\myim{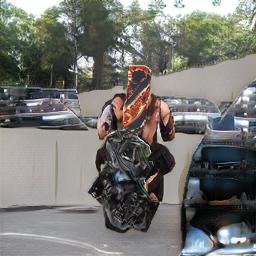}\\
\myim{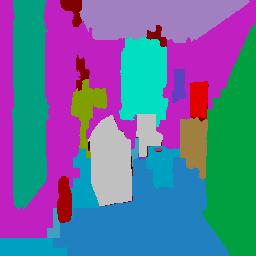} &
\myim{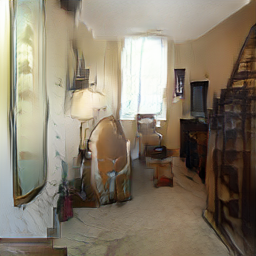} &
\myim{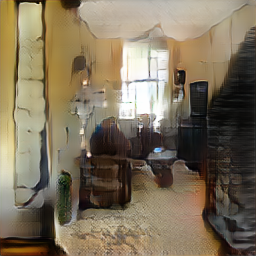}&
\myim{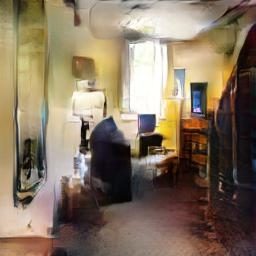} &
\myim{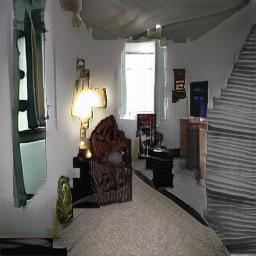}\\

     \end{tabular}
\label{fig:cocosemisup_supmat}
\end{figure*}

\begin{figure*}
\caption{Randomly selected samples of semantic image synthesis on Cityscapes25K in full data setting.
}
\def\myim#1{ \includegraphics[width=0.25\textwidth]{figures/#1}}
     \setlength\tabcolsep{0.5 pt}
   \renewcommand{\arraystretch}{0.2}
   \vspace{0.3cm}
    \centering
   \begin{tabular}{cccccc}
   &SPADE & OASIS&  \ours\\
\myim{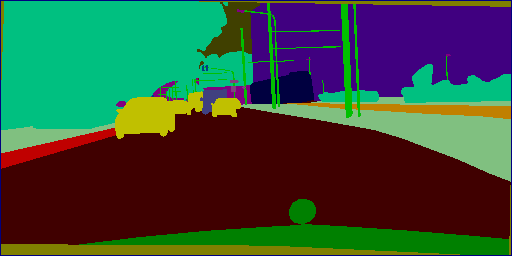} &\myim{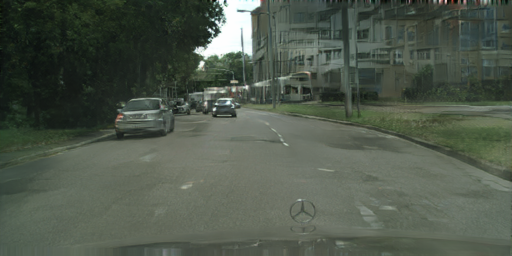} &
\myim{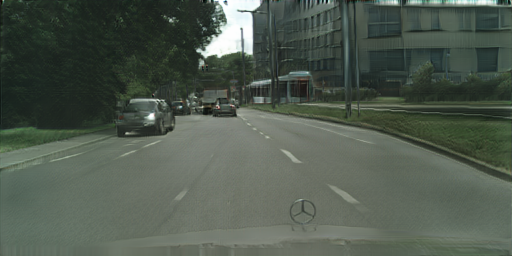} &
\myim{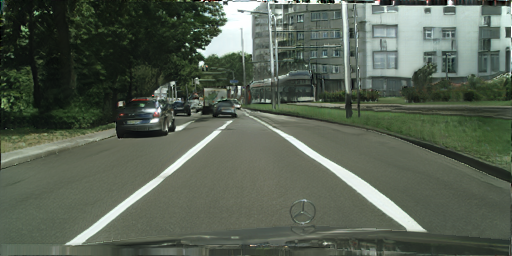} \\
\myim{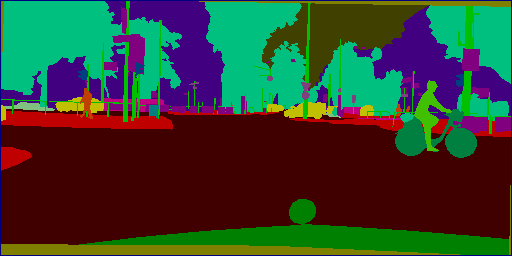} &
\myim{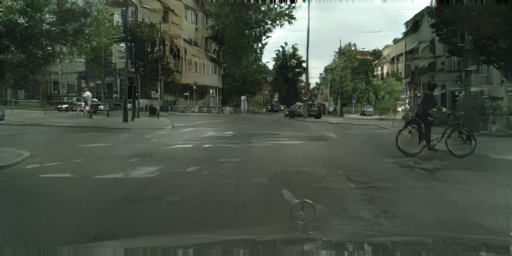} &
\myim{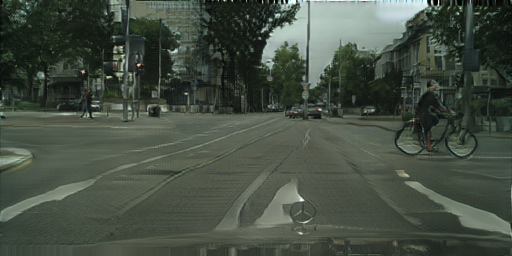} &
\myim{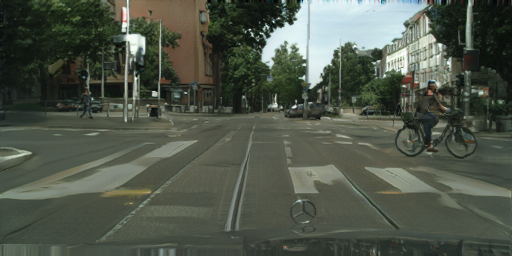} \\
\myim{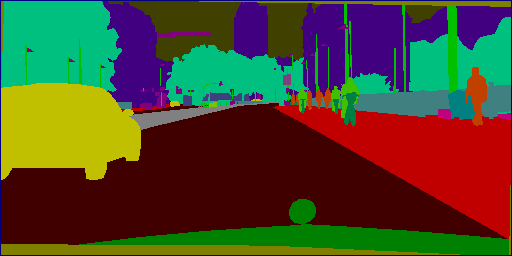} &
\myim{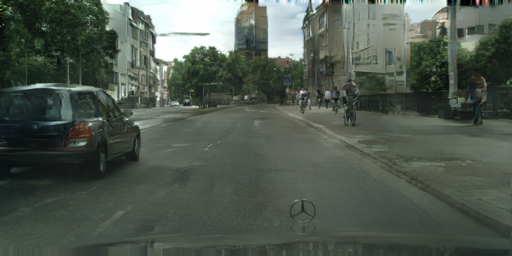} &
\myim{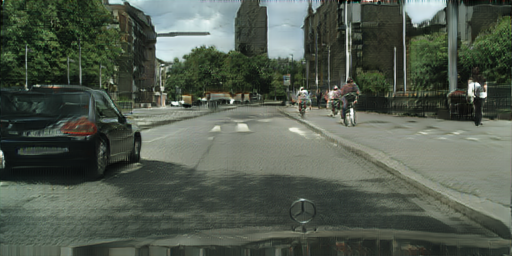} &
\myim{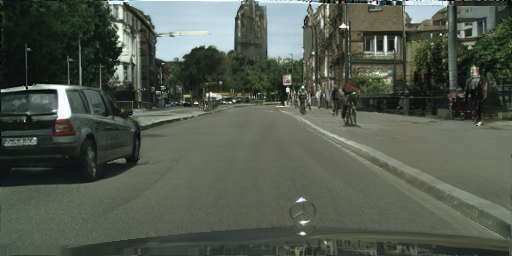} \\
\myim{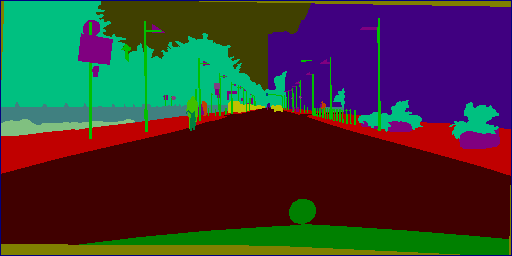} &
\myim{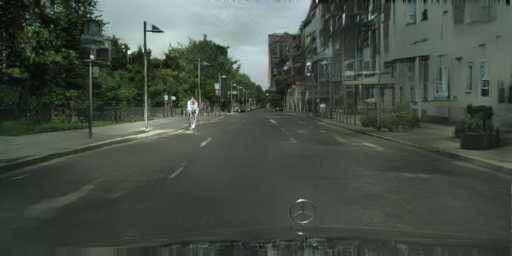} &
\myim{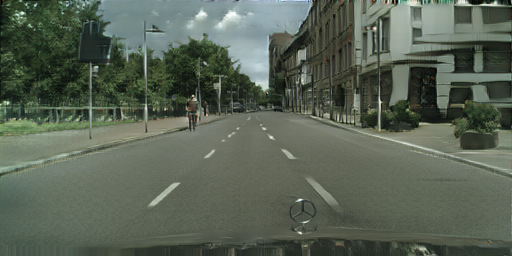} &
\myim{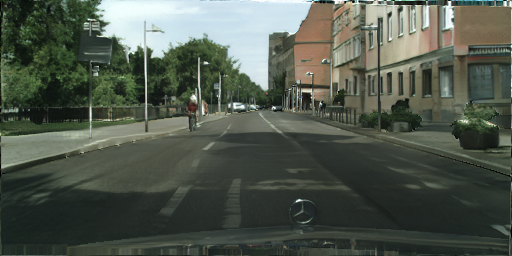} \\

\myim{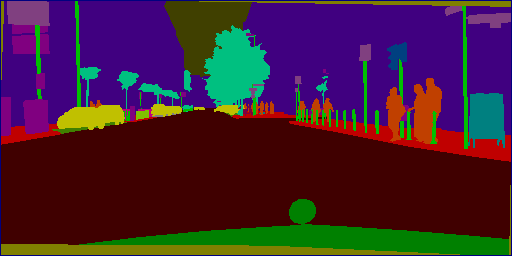} &
\myim{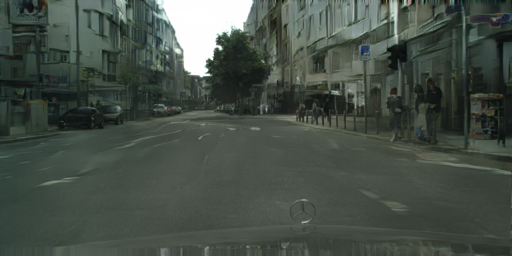} &
\myim{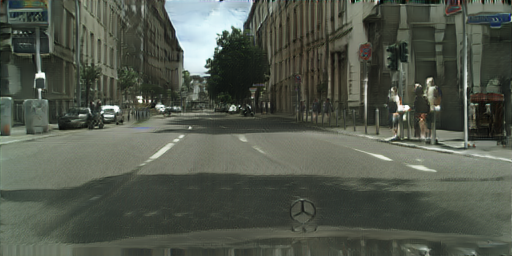} &
\myim{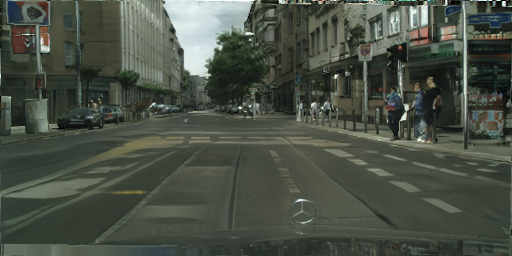} \\
\myim{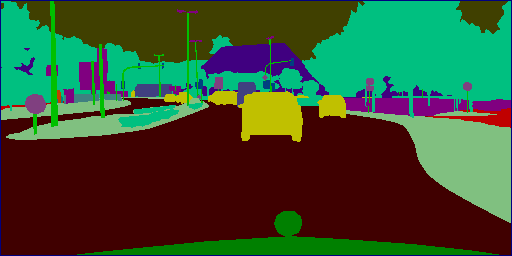} &
\myim{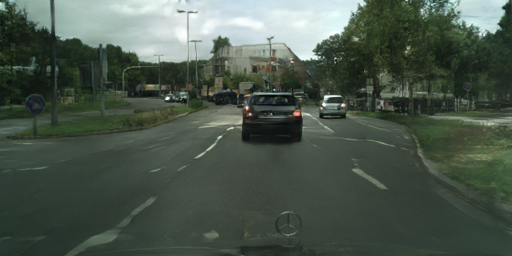} &
\myim{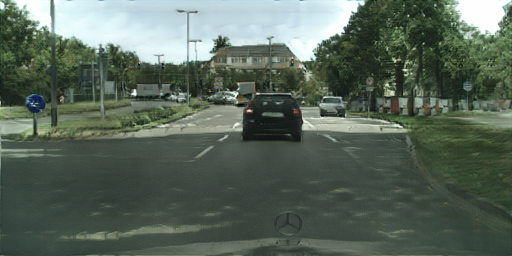} &
\myim{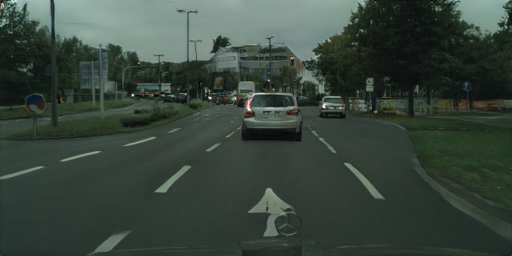} \\

\end{tabular}
\label{fig:cityscapescondmoredata}
\end{figure*}

\begin{figure*}
\caption{Randomly selected unconditional (left)  and conditional (right) samples for ADE20K in full data setting. }
 \def\myim#1{ \includegraphics[width=16.5mm,height=16.5mm]{figures/#1}}
 \vspace{0.3cm}
   \begin{minipage}[t]{0.37\textwidth}
   \setlength\tabcolsep{0.5pt}
   \renewcommand{\arraystretch}{0.2}
   \tiny
    \centering
     \begin{tabular}{ccc}
     StyleMapGAN&SB-GAN&\ours\\
\myim{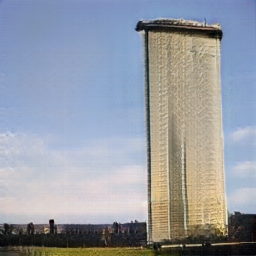} &
\myim{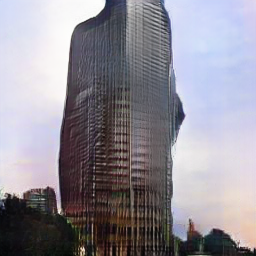} &
\myim{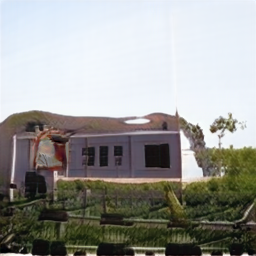}\\
\myim{ade20k_uncond/1_stylemapgan.png} &
\myim{ade20k_uncond/1_sbgan.png} &
\myim{ade20k_uncond/1_hybrid.png}\\
\myim{ade20k_uncond/2_stylemapgan.png} &
\myim{ade20k_uncond/2_sbgan.png} &
\myim{ade20k_uncond/2_hybrid.png}\\
\myim{ade20k_uncond/3_stylemapgan.png} &
\myim{ade20k_uncond/3_sbgan.png} &
\myim{ade20k_uncond/3_hybrid.png}\\
\myim{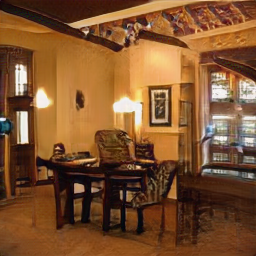} &
\myim{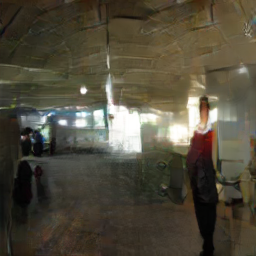} &
\myim{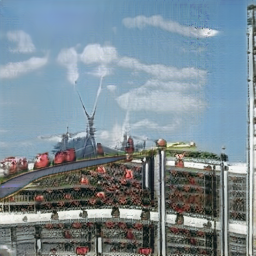}\\

      \end{tabular}
   \end{minipage}
   \hfill
   \begin{minipage}[t]{0.62\textwidth}
     \centering
   \setlength\tabcolsep{0.5 pt}
   \renewcommand{\arraystretch}{0.2}
   \tiny
     \begin{tabular}{ccccc}
      & OASIS& SBGAN & \ours\\
\myim{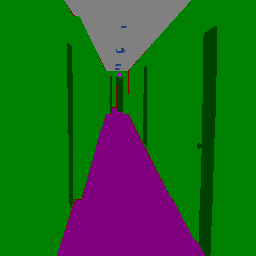} &

\myim{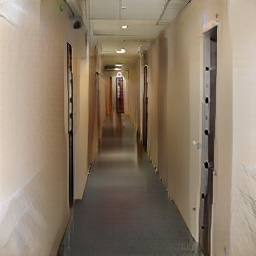} &
\myim{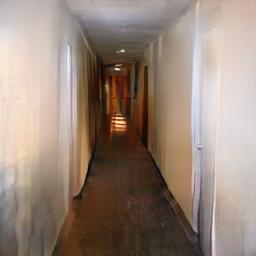} &
\myim{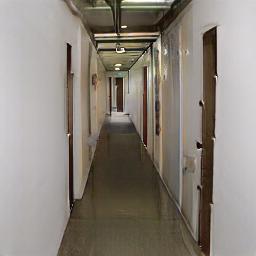}  \\
\myim{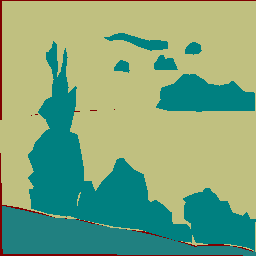} &

\myim{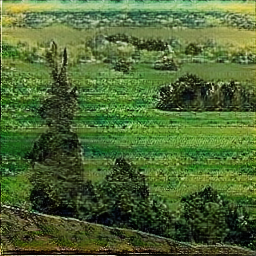} &
\myim{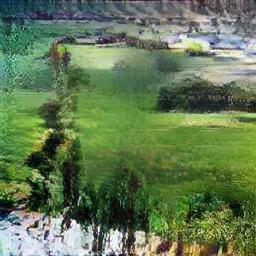} &
\myim{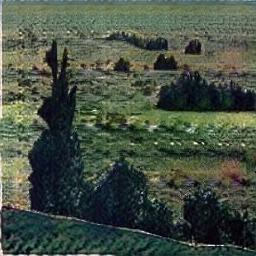}  \\
\myim{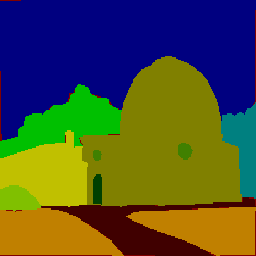} &

\myim{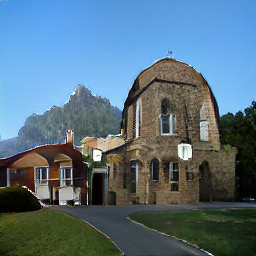} &
\myim{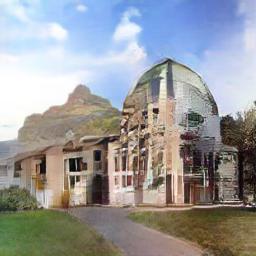} &
\myim{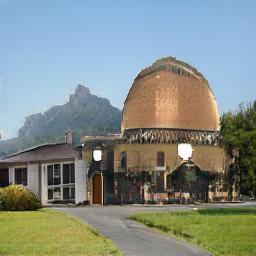}  \\
\myim{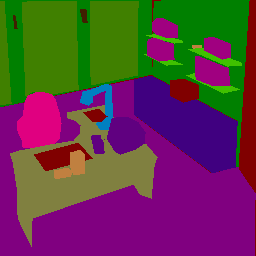} &

\myim{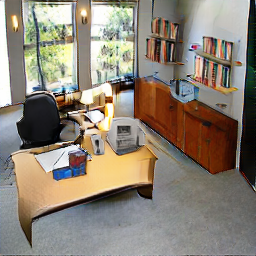} &
\myim{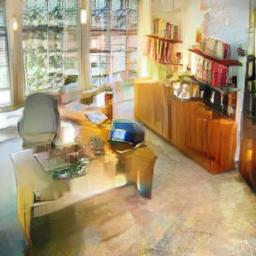} &
\myim{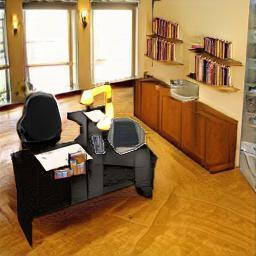}  \\
\myim{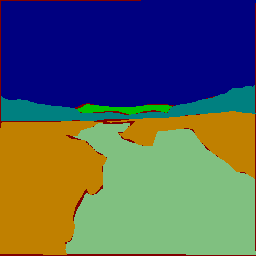} &

\myim{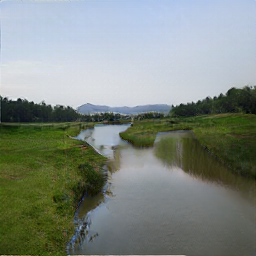} &
\myim{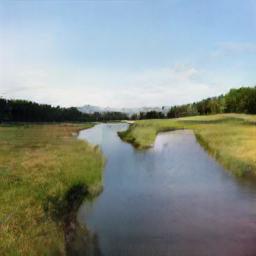} &
\myim{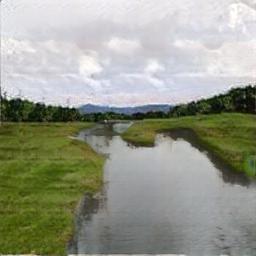}  \\

\end{tabular}
\end{minipage}    
\label{fig:ade20k_supmat}
\end{figure*}

\begin{figure*}
\caption{
Randomly selected unconditional (left)  and conditional (right) samples for COCO-Stuff in full data setting.  
}
 \def\myim#1{ \includegraphics[width=16.5mm,height=16.5mm]{figures/#1}}
 \vspace{0.3cm}
   \begin{minipage}[t]{0.37\textwidth}
   \setlength\tabcolsep{0.5pt}
   \renewcommand{\arraystretch}{0.2}
   \tiny
    \centering
     \begin{tabular}{ccc}
     StyleMapGAN&SB-GAN&\ours\\
\myim{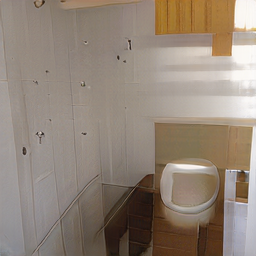} &
\myim{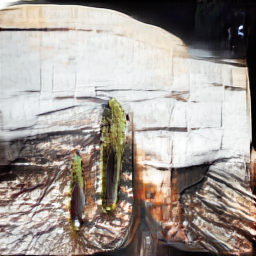}&
\myim{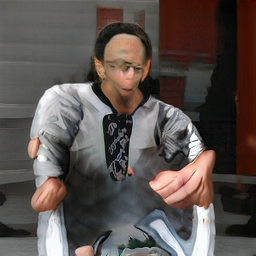}\\
\myim{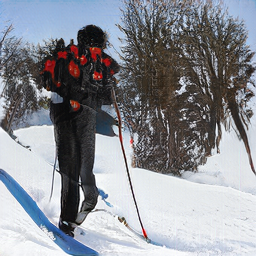} &
\myim{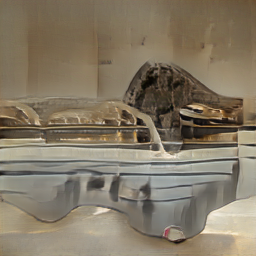}&
\myim{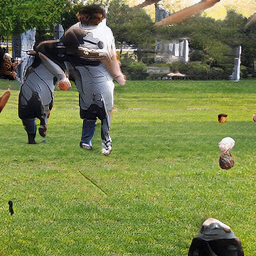}\\
\myim{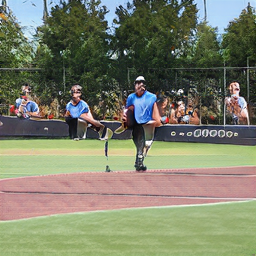} &
\myim{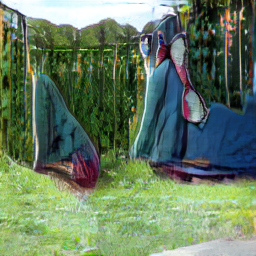}&
\myim{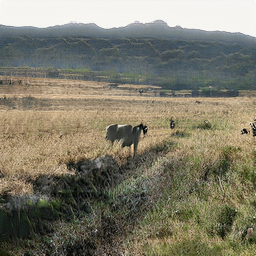}\\
\myim{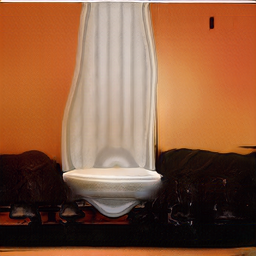} &
\myim{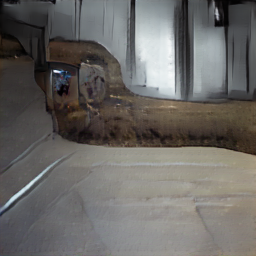}&
\myim{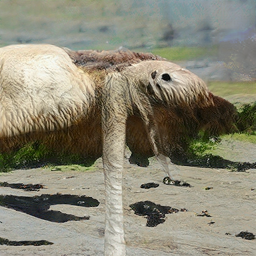}\\
\myim{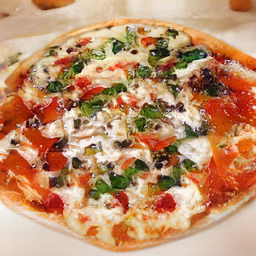} &
\myim{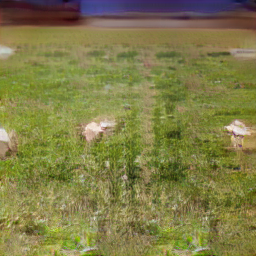}&
\myim{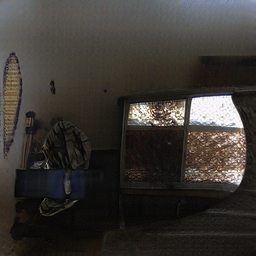}\\

      \end{tabular}
   \end{minipage}
   \hfill
   \begin{minipage}[t]{0.62\textwidth}
     \centering
   \setlength\tabcolsep{0.5 pt}
   \renewcommand{\arraystretch}{0.2}
   \tiny
     \begin{tabular}{ccccc}
     & OASIS& SBGAN & \ours\\
\myim{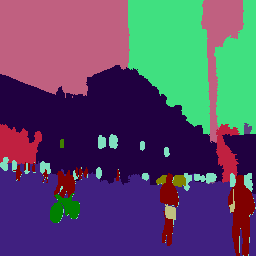} &

\myim{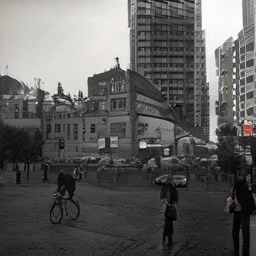}&
\myim{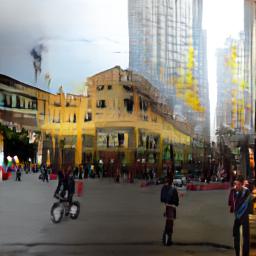} &
\myim{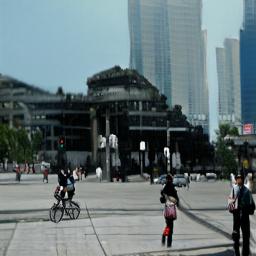}\\
\myim{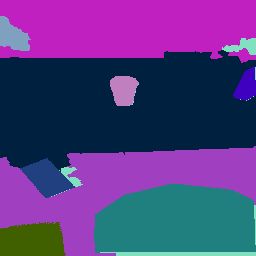} &

\myim{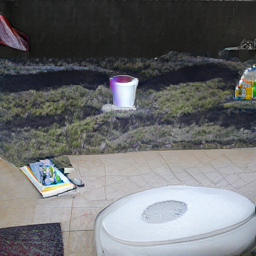}&
\myim{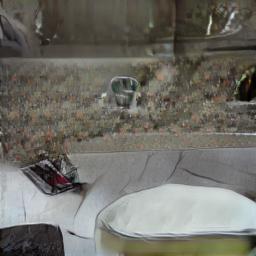} &
\myim{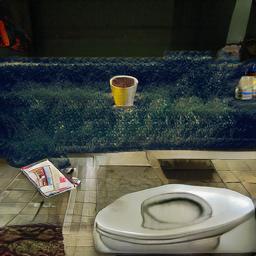}\\
\myim{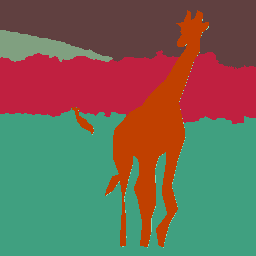} &

\myim{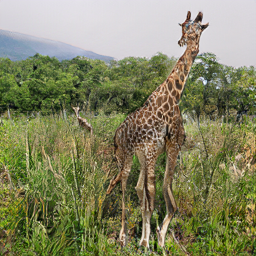}&
 \myim{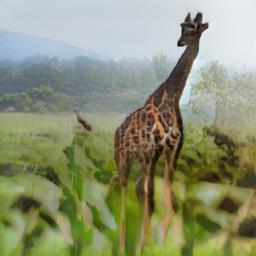}&
\myim{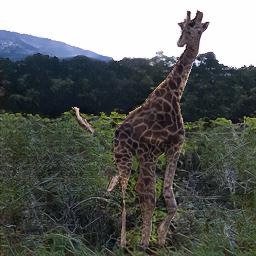}\\
\myim{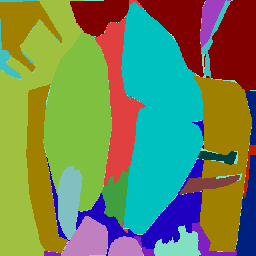} &

\myim{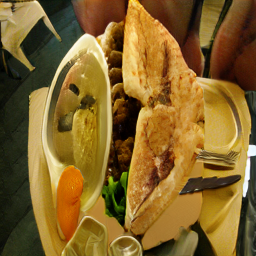}&
 \myim{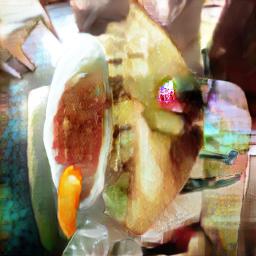}&
\myim{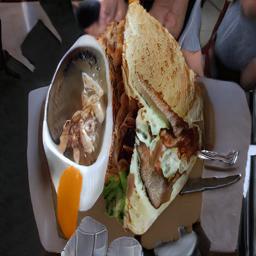}\\
\myim{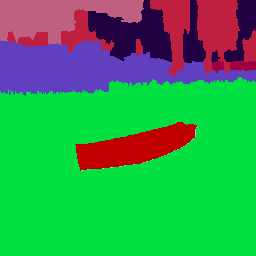} &

\myim{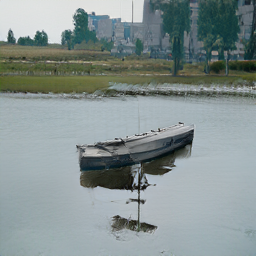}&
\myim{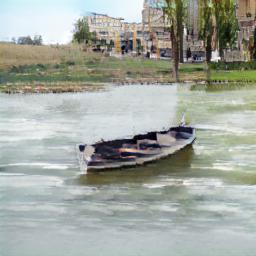} &
\myim{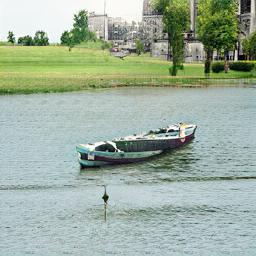}\\

     \end{tabular}
\end{minipage}    

\label{fig:coco_supmat}
\end{figure*}

\section{Additional quantitative results}
\label{sec:quantitative}

\begin{figure*}
\caption{Diversity generation on ADE20K, (several samples with same input conditioning) in full data setting}
 \setlength\tabcolsep{0.5 pt}
   \renewcommand{\arraystretch}{0.2}

\def\myim#1{ \includegraphics[width=100mm,height=20mm]{figures/#1}}
\def\myimone#1{ \includegraphics[width=20mm,height=20mm]{figures/#1}}
  \centering
    \begin{tabular}{ccc}
    && \ours\\
    &
     \myimone{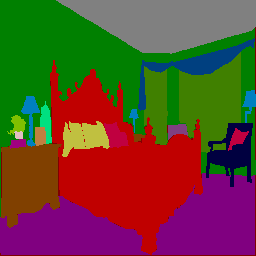}
    &
     \myim{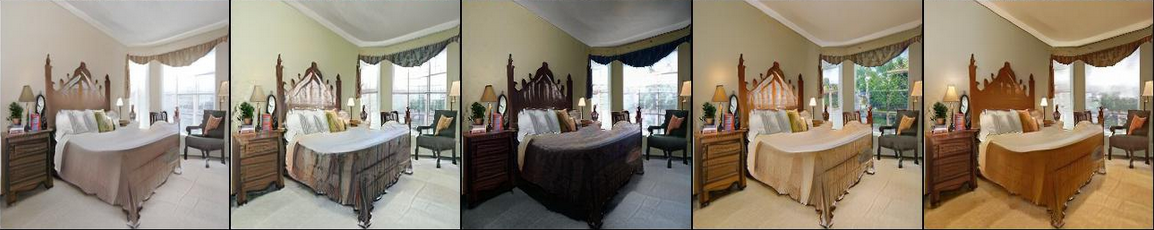} \\
     &
     \myimone{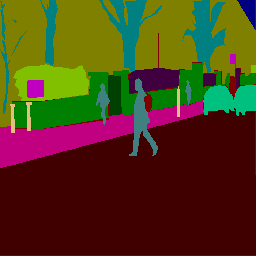}
     &
     \myim{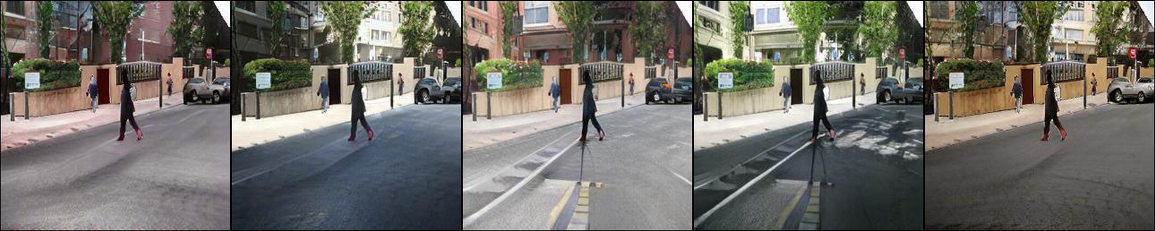} \\
      &
     \myimone{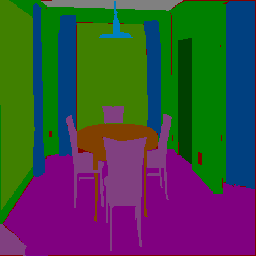} 
     &
     \myim{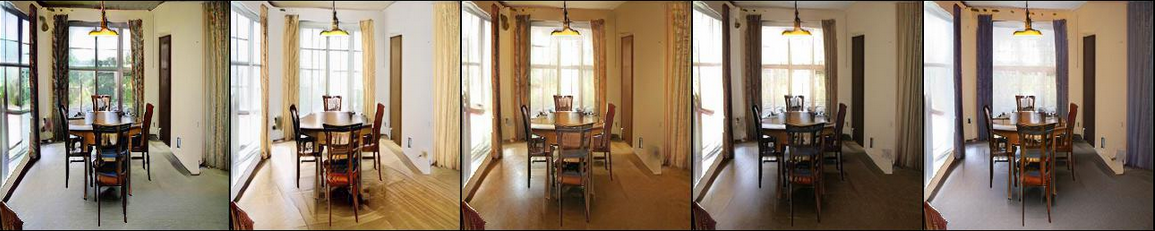} \\
      &
     \myimone{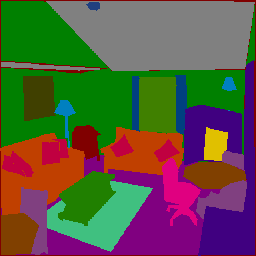}
     &
     \myim{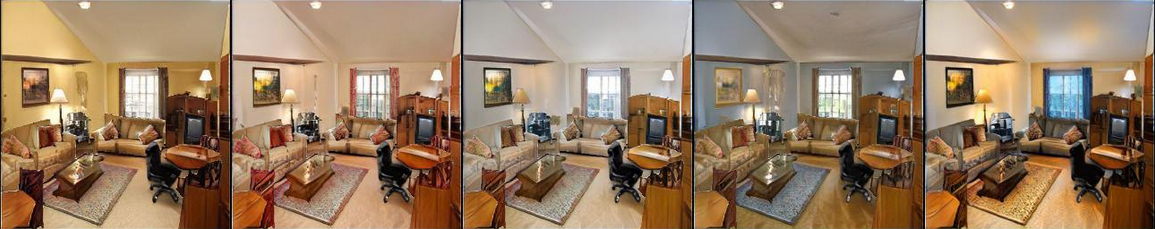} \\
     &
     \myimone{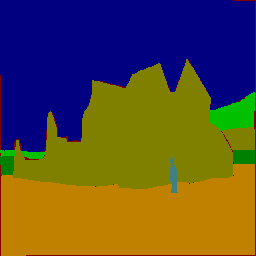}
     &
     \myim{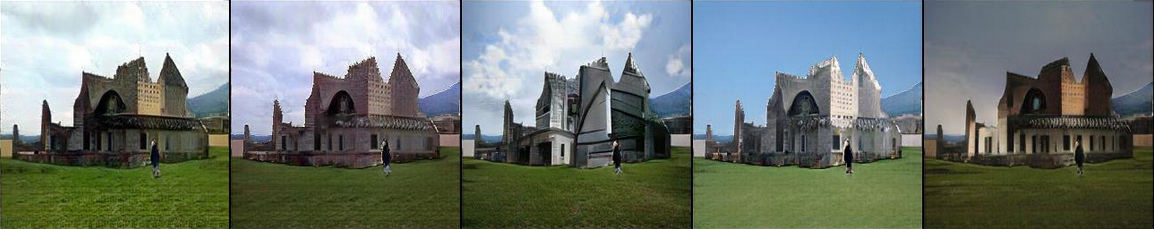} \\
    
     \end{tabular}

  l 
\label{fig:ade20kdiversity}
\end{figure*}

In Tables \ref{tab:smalldatamain_std_cityadeind}, \ref{tab:smalldatamain_std_ade4coco}, \ref{tab:supmat_semiannotatedAde20kCOCO}, \ref{table:fullAde20kCOCO_std}, \ref{tab:archi_supmat}, and \ref{tab:augment_std}  we provide the same results as in Tables~1,~2,~3,~4,~5 of the main paper, but add standard deviations obtained across five different sets of samples from the models.

 \begin{table*}
 \caption{Results in limited data setting for Cityscapes5K and ADE-Indoor, where OCO-GAN outperforms the leading semantic generative approach OASIS, as well as leading unconditional state-of-the-art StyleGAN2 and BigGAN. When available, results in italics are taken from original papers (in italic). In bold: Best result among  unconditional, conditional, and hybrid models, resp.
}
 \footnotesize
\centering
\resizebox{\textwidth}{!}{ 
 \begin{tabular}{@{}lccc|ccc@{}}
\toprule
  &  \multicolumn{3}{c}{\textbf{Cityscapes5K}} &  \multicolumn{3}{c}{\textbf{ADE-Indoor}} \\
  & 
 $\downarrow$\textbf{FID}  & 
 $\downarrow$\textbf{CFID} &   
 $\uparrow$\textbf{mIoU} &
 $\downarrow$\textbf{FID}  & 
 $\downarrow$\textbf{CFID} &   
 $\uparrow$\textbf{mIoU} \\  \midrule
StyleGAN2~\cite{Karras2019stylegan2} &  79.6 $\pm$ 3.1 &  & & 87.3 $\pm$ 3.3&  &  \\
StyleMapGan~\cite{kim21cvpr} &  \bf 70.4 $\pm$ 1.0 &  & & \bf 74.8 $\pm$ 1.2 & &  \\
BigGAN~\cite{brock19iclr} &  80.5 $\pm$ 1.5 &  & & 109.6 $\pm$ 1.4& &    \\
\midrule
SPADE~\cite{park19cvpr1} &  & \it 71.8 & \it62.3  &  & 50.3 $\pm$ 0.0 & 44.7 $\pm$ 0.0 \\
OASIS~\cite{sushko21iclr} &   &  \it \bf 47.7 & \it \bf 69.3  &  & \bf 50.8 $\pm$ 0.0 & {\bf57.5 $\pm$ 0.5}  \\
\midrule
PSP~\cite{richardson21cvpr} &  79.6 $\pm$ 3.1 & 118.7 $\pm$ 0.3 & 19.3 $\pm$ 0.1& 87.3 $\pm$ 3.3 & 148.7 $\pm$ 0.5& 35.6 $\pm$ 1 \\
SB-GAN~\cite{azadi2019semantic} & \it 65.5&   \it 60.4 &  n/a  &  \it 85.3 & \it 48.2 &  n/a \\
 \ours & {\bf 57.4 $\pm$ 1.3} &  {\bf 41.6 $\pm$ 0.1}& {\bf 69.4 $\pm$ 0.2} & {\bf69.0 $\pm$ 2} & {\bf47.9 $\pm$ 0.2} & \bf  58.6 $\pm$ 1.5\\
\bottomrule
\end{tabular}
}
 
\label{tab:smalldatamain_std_cityadeind}
\end{table*}

\begin{table*}
\caption{Results in limited data setting for ADE4K and COCO-Stuff12K, where OCO-GAN outperforms the leading semantic generative approach OASIS, as well as leading unconditional state-of-the-art StyleGAN2 and BigGAN. When available, results in italics are taken from original papers (in italic).  In bold: Best result among  unconditional, conditional, and hybrid models, resp.
}
 \footnotesize
\centering
\resizebox{\textwidth}{!}{ 
 \begin{tabular}{@{}lccc|ccc@{}}
\toprule
  &  \multicolumn{3}{c}{\textbf{ADE4K}} & \multicolumn{3}{c}{\textbf{COCO-Stuff12K}}\\
  & 
 $\downarrow$\textbf{FID}  & 
 $\downarrow$\textbf{CFID} &   
 $\uparrow$\textbf{mIoU} &
 $\downarrow$\textbf{FID}  & 
 $\downarrow$\textbf{CFID} &   
 $\uparrow$\textbf{mIoU} \\  \midrule
StyleGAN2~\cite{Karras2019stylegan2} &  95.7 $\pm$ 1.3 &  &  &  77.4 $\pm$ 0.7 &  &   \\
StyleMapGan~\cite{kim21cvpr} &  \bf 87.8 $\pm$ 0.7 &  &  &  \bf 65.2 $\pm$ 0.3 &   &   \\
BigGAN~\cite{brock19iclr} &   112.7 $\pm$ 0.9 &  &  &  82.2 $\pm$ 0.4 &  &  \\
\midrule

SPADE~\cite{park19cvpr1} &    & {\bf 43.4 $\pm$ 0.0}  & 34.3 $\pm$ 0.0 &  & 30.5 $\pm$ 0.0  & 33.8 $\pm$ 0.0\\
OASIS~\cite{sushko21iclr} &     & 49.9 $\pm$ 0.0 & {\bf36.1 $\pm$ 0.1} &   & {\bf 25.6 $\pm$ 0.0} & \bf 37.6 $\pm$ 0.1 \\
\midrule

PSP~\cite{richardson21cvpr} &  95.7 $\pm$ 1.3 & 157.7 $\pm$ 0.2 & 3.2 $\pm$ 0.6 & 77.4 $\pm$ 0.7 & 225.1 $\pm$ 0.8 & 3.1 $\pm$ 0.3\\
SB-GAN~\cite{azadi2019semantic} & 101.5 $\pm$ 0.2 & 49.3 $\pm$ 0.0 & 24.9 $\pm$ 0.2& 99.6 $\pm$ 0.4 & 52.6 $\pm$ 0.0 & 27.2  $\pm$ 0.2\\
 \ours & {\bf 80.6 $\pm$ 0.6} & \bf 43.8 $\pm$ 0.0  &  {\bf36.1 $\pm$ 0.3} & {\bf 51.5 $\pm$ 0.1}  & {\bf25.6 $\pm$ 0.0}  & {\bf40.3 $\pm$ 0.1}\\
\bottomrule
\end{tabular}
}
 
\label{tab:smalldatamain_std_ade4coco}
\end{table*}

\begin{table*}
\caption{Results in the  {\bf semi-supervised setting}. In bold: Best result among  unconditional, conditional, and hybrid models, resp.}
\footnotesize
\centering
\setlength{\tabcolsep}{1pt} 
\resizebox{\textwidth}{!}
{ 
 \begin{tabular}{lccc|ccc|ccc}
\toprule
   & \multicolumn{3}{c}{\textbf{Cityscapes}}& \multicolumn{3}{c}{\textbf{ADE20K}} & \multicolumn{3}{c}{\textbf{COCO-Stuff}}\\
& 
 $\downarrow$\textbf{FID}  &   $\downarrow$\textbf{CFID} &    $\uparrow$\textbf{mIoU}   &
 $\downarrow$\textbf{FID}  &   $\downarrow$\textbf{CFID} &    $\uparrow$\textbf{mIoU}   &
 $\downarrow$\textbf{FID}  &   $\downarrow$\textbf{CFID} &    $\uparrow$\textbf{mIoU}   \\ \midrule
StyleGAN2~\cite{Karras2019stylegan2} & 49.1 $\pm$ 0.7 & && 43.1  $\pm$ 0.3  &   &  &  37.0 $\pm$ 0.4 &   &   \\
StyleMapGAN~\cite{kim21cvpr} & {\bf48.8}  $\pm$ 0.1 & & & \bf 39.6  $\pm$ 0.2  &   &   & \bf 33.0  $\pm$ 0.1  &   &   \\
BigGAN~\cite{brock19iclr}  &  67.8 $\pm$ 0.4 & & & 81.0 $\pm$ 0.5 & & & 70.5 $\pm$ 0.1 &  &    \\
\midrule

SPADE~\cite{park19cvpr1} &    &   83.7 $\pm$ 0.0 & 52.1 $\pm$ 0.0  & & \bf 43.4  $\pm$ 0.0 & 34.3  $\pm$ 0.0 &  &  \bf 42.0 $\pm$ 0.0 & \bf 28.6 $\pm$ 0.0 \\
OASIS~\cite{sushko21iclr} &   & \bf 77.4 $\pm$ 0.2 &  \bf 63.5 $\pm$ 0.0 & & 49.9 $\pm$ 0.0 &  \bf 36.1  $\pm$ 0.1 &  & 77.2 $\pm$ 0.0 & 23.5 $\pm$ 0.0 \\
\midrule
SB-GAN~\cite{azadi2019semantic} & 96.1 $\pm$ 1.2 & 93.1 $\pm$ 0.0 & 50.2 $\pm$ 0.4& 82.6  $\pm$ 0.3 & 45.5 $\pm$  0.0  & 31.0 $\pm$  0.2 & 103.5 $\pm$ 0.3& 57.6 $\pm$ 0.0 & 19.2  $\pm$ 0.5 \\

PSP~\cite{richardson21cvpr} &  \bf 49.1 $\pm$ 0.7 & 111.8 $\pm$ 0.1 & 13.8 $\pm$ 0.1 & 43.1 $\pm$ 0.3 &  147.2 $\pm$ 0.2 & 2.8 $\pm$ 0.4& \bf 37.0 $\pm$ 0.4 & 168.1 $\pm$ 0.1 & 2.4 $\pm$ 0.3 \\

\ours & 49.8 $\pm$ 0.4 & \bf 55.8 $\pm$ 0.2 & \bf 65.2 $\pm$ 0.4 & \bf 40.4 $\pm$ 0.8 & \bf 40.1 $\pm$ 0.0& \bf 37.6 $\pm$ 0.6 & 44.4 $\pm$ 0.2 & \bf 39.0 $\pm$ 0.0& \bf 32.1 $\pm$ 0.0\\

\bottomrule
\end{tabular}
}

\label{tab:supmat_semiannotatedAde20kCOCO}
\end{table*}

  \begin{table*}
  \caption{Quantitative results on the full Cityscapes, ADE20K and COCO-Stuff datasets. Results for other methods obtained from models we trained using code released by authors, or taken from original papers when available (marked in italic). *: SB-GAN on Cityscapes uses additional semantic segmentations obtained using a segmentation network during training. In bold: Best result among  unconditional, conditional, and hybrid models, resp.
}
 \footnotesize
\centering

\setlength{\tabcolsep}{1pt} 
\resizebox{\textwidth}{!}
{\scriptsize
 \begin{tabular}{lccc|ccc|ccc}
\toprule
   & \multicolumn{3}{c}{\textbf{Cityscapes}}& \multicolumn{3}{c}{\textbf{ADE20K}} & \multicolumn{3}{c}{\textbf{COCO-Stuff}}\\
  & 
  $\downarrow$\textbf{FID}  &   $\downarrow$\textbf{CFID} &    $\uparrow$\textbf{mIoU}   &
 $\downarrow$\textbf{FID}  &   $\downarrow$\textbf{CFID} &    $\uparrow$\textbf{mIoU}   &
 $\downarrow$\textbf{FID}  &   $\downarrow$\textbf{CFID} &    $\uparrow$\textbf{mIoU}   \\  \midrule
Stylegan2~\cite{Karras2019stylegan2} & 49.1 $\pm$ 0.7&&& 43.1 $\pm$ 0.3 &   &   & \bf 37.0 $\pm$ 0.4 &   &   \\
StyleMapGAN~\cite{kim21cvpr} & \bf 48.8 $\pm$ 0.1 &&&\bf 39.6 $\pm$ 0.2 &   &   &  33.0 $\pm$ 0.1 &  & \\
BigGAN &  67.8 $\pm$ 0.4 &&&81.0 $\pm$ 0.5 &  & & 70.5 $\pm$ 0.1 & & \\
\midrule

SPADE~\cite{park19cvpr1} &  & \it 71.8& \it 62.3&& \it 33.9 &  \it 38.5  &  & \it 22.6 & \it 37.4 \\
OASIS~\cite{sushko21iclr} & &\it \bf  47.7&  \textbf{\emph{ 69.3}} &&  \textbf{\emph{28.3}} &  \textbf{\emph{ 48.8}}  & & \textbf{\emph{ 17.0 }}& \textbf{\emph{ 44.1}}\\
\midrule

SB-GAN~\cite{azadi2019semantic} & \it  63.0 & \it 54.1&  n/a & 62.8 $\pm$ 0.1 & 35.0 $\pm$ 0.0 & 34.5 $\pm$ 0.2& 92.1 $\pm$ 0.7 &  54.7 $\pm$ 0.0 & 19.3 $\pm$ 0.1 \\
PSP~\cite{richardson21cvpr} & 49.1 $\pm$ 0.7 & 99.2 & 15.6 $\pm$ 0.2 & 43.1 $\pm$ 0.3 & 146.9 &2.8 $\pm$ 0.4 &\bf  37.0 $\pm$ 0.4 & 211.6 & 2.1 $\pm$ 0.6\\
\ours &  {\bf48.8 $\pm$ 0.2} &{\bf 41.5 $\pm$ 0.1 } & \bf 69.1 $\pm$ 0.1& {\bf38.6 $\pm$ 0.1} & \bf 30.4 $\pm$ 0.0 & \bf 47.1 $\pm$ 0.2 & 38.5 $\pm$ 0.1 & \bf 20.9 $\pm$ 0.0 &  \bf 42.1 $\pm$ 0.1\\
\bottomrule
\end{tabular}
}
\label{table:fullAde20kCOCO_std}
\end{table*}

\begin{table*}[t]
\caption{Ablation of architectures and training modes. We evaluate models in the limited data  setting with Cityscapes5K and semi-supervised setting with Cityscapes25K (with 500 annotated images).
Italics: results  taken from original paper. 
Best result in each column is printed in bold.}
\centering
\setlength{\tabcolsep}{1pt} 

\resizebox{\textwidth}{!}
{\small
 \begin{tabular}{lccc|ccc}
 
\toprule
  &  \multicolumn{3}{c|}{\textbf{Cityscapes5K}} &  \multicolumn{3}{c}{\textbf{Cityscapes25K}} \\
 & 
$\downarrow$\textbf{FID}  &   $\downarrow$\textbf{CFID} &    $\uparrow$\textbf{mIoU} &    
$\downarrow$\textbf{FID}  &   $\downarrow$\textbf{CFID} &    $\uparrow$\textbf{mIoU}   \\ \midrule

StyleGAN2~\cite{Karras2019stylegan2}  &  79.6 $\pm$ 3.1  &   &    &   49.1 $\pm$ 0.7 &   &   \\
StyleMapGAN~\cite{kim21cvpr} &   70.4  $\pm$ 1.0  &   &  &  \bf 48.8  $\pm$ 0.1 &   &  \\
\ours{ }- uncond.\  only  &  82.0 $\pm$ 1.8 &   &  &  50.5  $\pm$ 0.6&   &   \\
\midrule
OASIS~\cite{sushko21iclr} &   &  \emph 47.7   & \emph 69.3  &   &  77.4  $\pm$ 0.2  &   63.5 $\pm$ 0.0 \\
 \ours{ }- cond.\ only &   &  43.7 $\pm$ 0.2 & \bf 70.3   $\pm$ 0.2 &   &  76.8  $\pm$ 0.2 & 57.7 $\pm$ 0.2 \\
\midrule
\ours & {\bf57.4 $\pm$ 1.3} & {\bf41.6 $\pm$ 0.1} & 69.4 $\pm$ 0.2 & 49.8 $\pm$ 0.4 & \bf 55.8 $\pm$ 0.2& \bf 65.2 $\pm$ 0.4\\
\bottomrule
\end{tabular}
}

\label{tab:archi_supmat}
\end{table*}

\begin{table*}
\caption{
FID of StyleGAN2 and \ours (trained in hybrid mode) with and without Adaptive Data Augmentation (ADA). 
}
 \small
\centering
 \begin{tabular}{lrrrr}
\toprule
&  \textbf{City5K} & \textbf{ADE-Ind} & \textbf{ADE4K}  & \textbf{CCS12K}\\ 
\midrule
StyleGAN2~\cite{Karras2019stylegan2}    & 79.6 $\pm$ 3.1 &  87.3 $\pm$ 3.3  & 95.7 $\pm$ 1.3  & 77.4 $\pm$ 0.7 \\
same w/ ADA & 60.6 $\pm$ 1.0 &   63.9 $\pm$ 0.8  & 77.8  $\pm$ 0.3 & \bf48.9 $\pm$ 0.1 \\
\midrule
\ours       & 57.4 $\pm$ 1.3  &  69.0 $\pm$ 2.0 & 80.6 $\pm$ 0.6  & 51.5 $\pm$ 0.1 \\
same w/ ADA &{\bf48.0 $\pm$ 0.7}  & \bf 61.2 $\pm$ 1.3 &\bf67.2 $\pm$ 0.3 &50.0 $\pm$ 0.1\\
\bottomrule
\end{tabular}
\vspace{-1ex}

\label{tab:augment_std}
 \end{table*}

\end{document}